%% file: main.tex
\documentclass[review]{elsarticle}
\usepackage{times}
\usepackage{epsfig}
\usepackage{graphicx}
\usepackage{amsmath}
\usepackage{amssymb}

\usepackage{color}
\usepackage{algorithm}
\usepackage{algpseudocode}
\usepackage{dsfont}
\usepackage{subcaption}
\usepackage{booktabs}
\usepackage{tabularx}
\usepackage{url}
\usepackage{soul}
\usepackage{balance}
\usepackage{multirow}

\usepackage{lscape} 
\usepackage{array,booktabs}

\input{defs}

\usepackage[pagebackref=false,breaklinks=true,letterpaper=true,colorlinks,bookmarks=false]{hyperref}

\journal{Journal of Pattern Recognition}

\begin{document}

\begin{frontmatter}

\title{Repurposing Existing Deep Networks for Caption and Aesthetic-Guided Image Cropping} %

\author[a]{Nora Horanyi}
\author[b]{Kedi Xia}
\author[c]{Kwang Moo Yi}
\author[c]{Abhishake Kumar Bojja}
\author[a]{\\Ale\v{s} Leonardis}
\author[a]{Hyung Jin Chang\corref{mycorrespondingauthor}}
\ead{h.j.chang@bham.ac.uk}
\cortext[mycorrespondingauthor]{Corresponding author}

\address[a]{University of Birmingham, United Kingdom}
\address[b]{Zhejiang University, China}
\address[c]{University of Victoria, Canada}

\input{tex/abstract}

\begin{keyword}
\texttt{Image cropping, Aesthetics, Deep network re-purposing, Image captioning}
\end{keyword}

 \end{frontmatter}

 \input{tex/intro_hj}
 \input{tex/related}

 \input{tex/method}

 \input{tex/result}

 \input{tex/conclusion}

 \input{tex/acknowledge}

 \thispagestyle{empty}

\input{references.bbl}
\bibliography{references.bbl}

\end{document}

%% file: defs.tex
\newcommand{\comment}[1]{}

\newcommand{\bI}{\mathbf{I}}

\newcommand{\bS}{\mathbf{S}}

\newcommand{\by}[0]{\mathbf{y}}

\newcommand{\calN}{\mathcal{N}}
\newcommand{\calU}{\mathcal{U}}

\DeclareMathOperator*{\argmin}{arg\,min}

\newcommand{\fig}[1]{Fig.~\ref{fig:#1}}

\newcommand{\tbl}[1]{Table~\ref{tbl:#1}}

\newcommand{\algo}[1]{Alg.~\ref{alg:#1}}

\newcommand{\loss}{\mathcal{L}}

\newcommand{\sample}{\text{Sample}}
\newcommand{\resize}{\text{Resize}}

\definecolor{orange}{rgb}{1,0.5,0}
\definecolor{blue}{rgb}{0,0,0.6}
\definecolor{green}{rgb}{0,0.6,0}
\definecolor{purple}{rgb}{0.5,0,0.5}

\definecolor{color1}{RGB}{0,199,1}
\definecolor{color2}{RGB}{224,43,28}

\newcommand{\eq}[1]{Eq.~\eqref{eq:#1}}

\newcommand{\btheta}{\boldsymbol{\theta}}

\newcommand{\etal}{\textit{et al}.}

\newcommand{\wrt}{\textit{w}.\textit{r}.\textit{t}.}

\newcolumntype{P}[1]{>{\centering\arraybackslash}p{#1}}

\renewcommand{\paragraph}[1]{\vspace{0.5em}\noindent\textbf{#1}}

%% file: tex/abstract.tex
\begin{abstract}
We propose a novel optimization framework that crops a given image based on user description and aesthetics.
Unlike existing image cropping methods, where one typically trains a deep network to regress to crop parameters or cropping actions, we propose to {\em directly} optimize for the cropping parameters by repurposing pre-trained networks on image captioning and aesthetic tasks, without any fine-tuning, thereby avoiding training a separate network.
Specifically, we search for the best crop parameters that minimize a combined loss of the initial objectives of these networks.
To make the optimization stable, we propose three strategies: 
(i) multi-scale bilinear sampling, (ii) annealing the scale of the crop region, therefore effectively reducing the parameter space, (iii) aggregation of multiple optimization results.
Through various quantitative and qualitative evaluations, we show that our framework can produce crops that are well-aligned to intended user descriptions and aesthetically pleasing.
\end{abstract}

%% file: tex/intro_hj.tex
\section{Introduction}
With the advent of social networks, it is now common that images are provided with captions and tags -- for example via Instagram or Twitter -- where captions themselves are highly tied in with the user's intentions regarding these images.
Therefore, an automated process for enhancing images, for example providing artistic crops or making thumbnail images that respect user intentions, would be useful for these social networking platforms (SNS).
Besides SNS applications, more tied in with  computer vision applications such as semi-automated image dataset generation \cite{Huang2007GeneratingGT}, tracking target object initialization \cite{iswanto2017visual}, story-based automatic image transition generation \cite{chu2014optimized}, can also benefit from text-based image cropping.
In this paper, as illustrated in \fig{teaser}, we focus on a novel image cropping task, which aims to crop an image automatically based on user intent -- expressed through a natural text-based description
-- and the aesthetics of the cropping outcome. 
\input{fig/teaser/item.tex}

Because of the usefulness of an automated image cropping system, various methods have been suggested.
However, existing image cropping methods~\cite{ChenCVPR2016,kao2017automatic,cornia2018automatic, guo2018} are typically designed to be purely automatic, leaving the user out of the loop. 
For example, \cite{cornia2018automatic} automatic cropping is based on the maximization of the saliency inside the cropping region. Attention-based methods like this try to preserve the most salient part of the image during cropping. One major downside of them is that the user cannot influence their behaviour.
Recent works~\cite{shan2018photobomb,li2018a2,chen2017quantitative,ChenACM2017} have focused on making this process even easier through automatically cropping a photo-based on aesthetics.
Despite being fully automated, aesthetics-based methods leave no room for the user to intervene.
Furthermore, 
they
provide no guarantee that the initial content and the intent of the images are preserved.

Efforts have also been taken toward methods that take user intention as input.
Description-based object detection~\cite{mattnet} and localization \cite{grounding2016} have recently been proposed for this purpose.
However, these description-based methods do not consider how natural images are created, and their behaviours are far from how humans would crop.
Most distinctively, these methods provide very tight cropping around an object.

As in many other areas of computer vision,
towards this goal, one could apply an end-to-end deep learning framework to find crops that fit the descriptions given an image~\cite{mattnet}.
However, training such a network in a typical supervised deep learning setup would require an immense amount of labelled data, with captions for multiple sub-regions of the images, as well as their respective \emph{``ground-truth''} crops, which may be subjective depending on the creator of the dataset. 
This is a challenging task.

Therefore, in this paper, we take an alternative approach that exploits existing networks trained for related tasks -- image captioning and aesthetics estimation -- and repurpose them to automatically crop the images, thereby avoiding the hardships of training a separate network.

Our contributions are four-fold:

\begin{itemize}
    \setlength{\itemsep}{0pt}%
    \setlength{\parskip}{0pt}%
    \vspace{-.5em}
    \item We propose a new deep networks repurposing framework to optimize crop parameters {\em directly} using a bilinear sampler \cite{Jaderberg15}, a pre-trained image captioning network \cite{xu2015show}, and a pre-trained aesthetic estimation network \cite{ChenACM2017}.
    \item We optimize to find the crop region that best fits the provided caption in terms of the image captioning network losses, as well as maximizes the aesthetics network scores.
    \item We generate a new dataset with multiple ground truth bounding box annotations for each caption. 
    \item With approaches above, we were able to not only outperform state-of-the-art methods but also produce more visually pleasing image crops with well-reflecting user's intention                                  
\end{itemize}

Utilizing the two networks requires special attention.
Since we are, in fact, repurposing the two pre-trained networks for a different purpose, we keep them intact to minimize the change inflicted upon them, and
propose a new loss term that we minimize instead of their original ones: learning to generate image captions and measuring how much image looks good.
For the image captioning network we propose to ignore the order of the words as we simply want the contents to be accurate.
For the aesthetic network, we directly try to maximize the aesthetic score.

Since this optimization process is, in its basic form, highly unstable due to the nature of the image gradients, we further propose a new optimization strategy based on \emph{scale annealing} and \emph{multiple restarts}.
Instead of directly optimizing for the position and the scale of the crop, we optimize only for the position and anneal the scale
throughout the optimization.
We use a \emph{multiple restart} technique, where for each scale we start from a random location and use the average of the optimization outcomes.
We further take advantage of the fact that image captions should not differ drastically as 
the image is blurred, 
and use
a multi-scale representation of the image for the captioning network.

%% file: fig/teaser/item.tex
\begin{figure}
\centering
\includegraphics[width=0.65\linewidth]{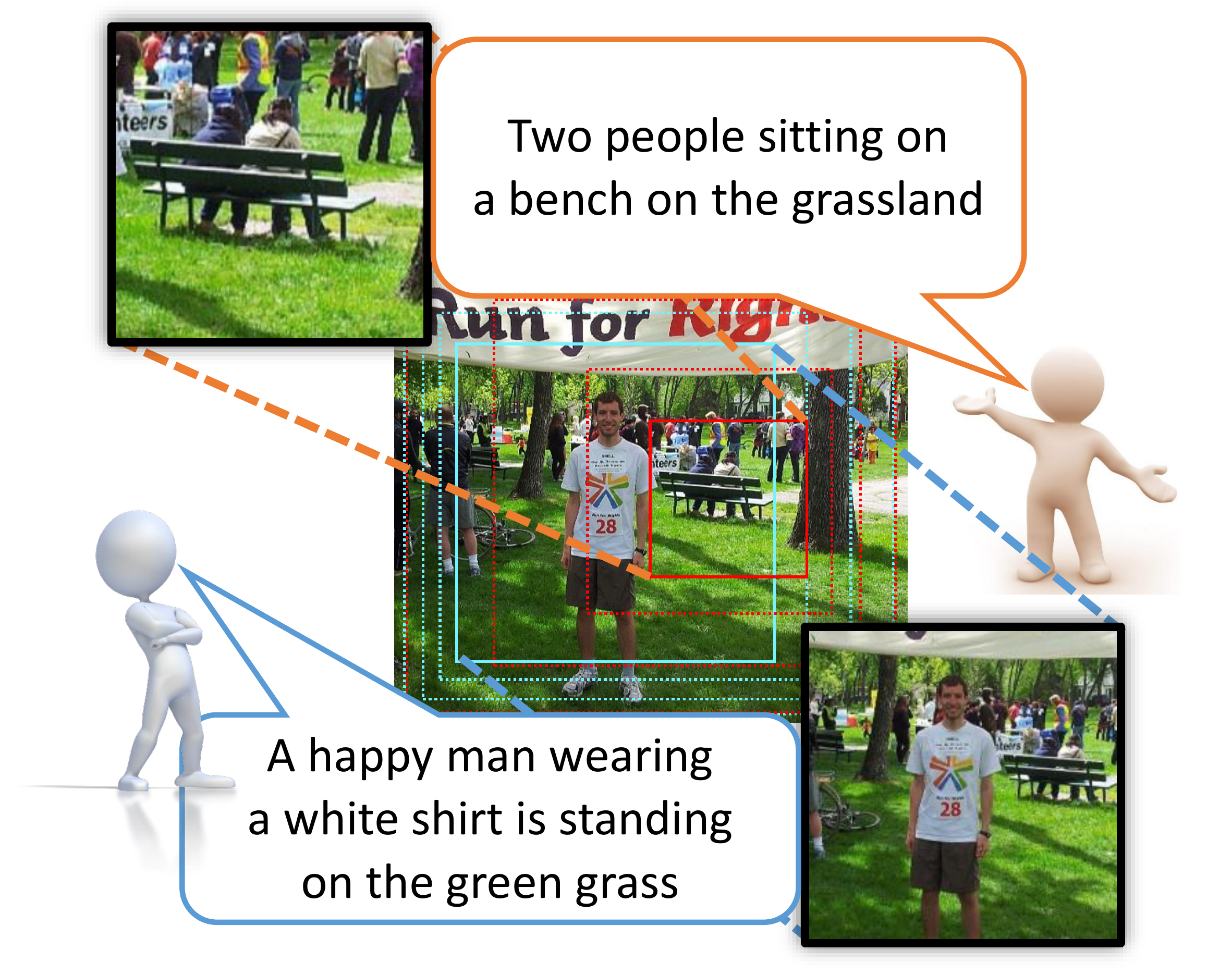}
\caption{Our proposed framework reliably crops visually pleasing images by following natural descriptions of users. This can produce different crops on the same image based on different users' descriptions.}
\vspace{-0.3cm}
\label{fig:teaser}
\end{figure}

%% file: tex/related.tex
\section{Related Works}

The image cropping research can be categorized into \emph{attention-based}, \emph{aesthetics-based}, and \emph{description-based} approaches.

\paragraph{Attention-based methods.} This class of methods exploits visual saliency models or salient object detectors to find the most visually important regions in the original image ~\cite{huang2015automatic,jaiswal2015saliency}. The majority of the attention-based cropping methods rank candidates based on their attention score ~\cite{choi2016object, FangACM2014}. Thus these methods are able to identify the most important and attractive regions of the image.    
Recent methods choose the important area based on certain attention scores~\cite{wang2015saliency, wang2015consistent, wang2016correspondence, wang2018saliency}. One of the most recent methods~\cite{cornia2018automatic}
focuses on those image regions which most probably attract the human gaze at first sight.

\paragraph{Aesthetics-based methods.} Methods based on aesthetics crop images by relying on the attractiveness of the cropped image with the help of a quality classifier ~\cite{ChenCVPR2016, islam2017survey,lu2019aesthetic}. The goal of these methods is to extract the optimal rectangular sub-region of a given image to produce an image with a high aesthetic score  ~\cite{guo2018,chen2017quantitative}. Recently, Li \etal~\cite{li2018a2} formulated the automatic image cropping problem as a sequential decision-making process and proposed a weekly supervised approach which only uses aesthetic information as supervision. This method was the first to use reinforcement learning for automatic image cropping and was able to overcome the disadvantages of the sliding window method.
The existing aesthetics-based cropping methods can be further categorized into supervised \cite{hong2017cnn,wang2015learning} or weekly supervised~\cite{kao2017automatic, ChenACM2017,FangACM2014,zhang2014weakly} methods. 
In particular, the weekly supervised methods, which do not include bounding box supervision, have been researched actively as producing cropping box annotations for the training is expensive \cite{goodview2018}.

\paragraph{Description-based methods.}
These methods aim to localize a region described by a given referring expression.
Most of these methods treat comprehension as bounding box localization, which is similar to our cropping task.
A recent method by Rohrbach \etal~\cite{grounding2016} uses joint embedding to find the object directly by selecting the best region based on an input expression.
Yu \etal~\cite{mattnet} proposed a modular network for referring expression comprehension. 
Similarly to our application, MAttNet \cite{mattnet} and Align2Ground \cite{datta2019align2ground} are able to locate the image region described by a general referring expression.
Their downfall is that these methods rely strictly on the input expressions and do not consider aesthetics at all. 
In addition, MAttNet heavily relies on three specific decompositions of the expression -- subject appearance, location, and relationship to other objects -- these may not exist in natural descriptions of images.

%% file: tex/method.tex
\section{Methodology}

We currently have deep networks designed to do many traditional computer vision tasks, and they perform in many cases even better than the traditional ones.
It is also quite common to use pre-trained networks as backbones for performing a certain task.
Here, we are proposing an alternative, and use these existing networks as building blocks.
We are proposing to go beyond the paradigm that deep networks should give a solution in a single shot, and instead perform inference through optimization, as was a common strategy before deep learning.

We first describe the overall architecture, including our \emph{multi-scale sampling strategy}, then explain how we perform inference by {\em optimizing} the framework instead of training networks to obtain the desired crop region.
We further detail how we can stabilize this optimization process through \emph{scale annealing} and \emph{multiple restarts}.

\subsection{Framework}
\input{fig/framework/item.tex}
\fig{framework} shows the overall framework of the proposed method.
Our framework comprises three major components: a) the bilinear sampler that operates on a multi-scale, b) the image captioning network, and c) the aesthetic network.

The image captioning network automatically generates a natural language expression describing the given image content. There is a large body of work on this problem \cite{bai2018survey}. Among those for the image captioning network, we use the method from \cite{xu2015show}, with the models pre-trained with the MS-COCO~\cite{Lin2014MicrosoftCC} dataset.
Visual aesthetic preference can be described either as a single score or as a distribution of scores. We follow the definition in \cite{ChenACM2017}, where the aesthetic score is provided by professional photographs.
For the aesthetic network we use \cite{ChenACM2017} with the pre-trained models\footnote{\scriptsize{\url{https://github.com/yiling-chen/view-finding-network}}}.

Note that we take special care that none of the images that were used in the training of any of the pre-trained models is included in our evaluation later.
\textit{Despite the fact that these two networks were trained on entirely different datasets, we found that the pre-trained models are good enough for our purpose.}

\paragraph{Multi-scale bilinear sampling. }
As we are directly optimizing for the location and scale of the crop region, it is essential that the gradients that the bilinear sampling gives are robust.
To ensure this, we propose to use a multi-scale strategy, inspired by the observation that, even when an image becomes blurry, its content does not change.
Therefore, if we denote the bilinear sampling process as $\sample\left(\bI,\btheta\right)$, where $\bI$ is the image and $\btheta$ is the crop parameters composed of the center coordinates of the crop $x$ and $y$, and its scale $s$, and $\resize\left(\cdot\right)$ is the resizing operation, for the cropped image  $\bI_{crop}\left(\btheta_s\right)$ we can write 
\begin{equation}
    \bI_{crop}\left(\btheta_s\right) = \frac{1}{\left|\bS\right|}\sum_{s\in\bS}\sample\left(\resize\left(\bI, s\right), \btheta\right) 
    ,
    \label{eq:crop}
\end{equation}
where $\bS \in \left\{\frac{1}{4},\frac{1}{3}, \frac{1}{2}, 1\right\}$ is the set of scales, and $\left|\bS\right|$ is the cardinality of this set, which is four. 
We also omit $\bI$ on the left-hand side for brevity.
Here, we set the sampling process to always consider the source image coordinates to be between $-1$ and $1$, thus removing the need of adjusting the sampling parameters per image. After each resizing operation, we also apply Gaussian image blurring filters.

\subsection{Inference}

With the framework we infer the parameters of the crop $\btheta$
by optimizing the network \wrt the image captioning networks loss and the aesthetic network output. 
However, as we are not using the two networks with their original purpose---learning to generate image captions or compare two images to find the one that looks better---, here, we formally introduce our optimization objective.

If we denote the two objectives as $\loss_{caption}$ for the image captioning part and $\loss_{aesthetic}$ for the aesthetic part, for a given pair of image $\bI$ and caption $\by$, our objective is therefore to find $\hat{\btheta}$ such that
\begin{equation}
    \hat{\btheta}  = 
    \argmin_{\btheta} \loss_{total}\left(\bI, \by, \btheta\right)
    ,
    \label{eq:objective}
    \vspace{-0.4cm}
\end{equation}
where
\begin{equation}
    \loss_{total}\left(\bI, \by, \btheta\right) = 
    \loss_{caption}\left(\bI, \by, \btheta\right) + \lambda \loss_{aesthetic}\left(\bI, \btheta\right)
    ,
    \label{eq:total}
\end{equation}
and $\lambda$ is the hyper-parameter that balances the two loss terms.
In all our experiments, we empirically set $\lambda=0.01$.
The two losses we choose to optimize are closely related to the networks' original formulation but modified to our need.

\paragraph{Image Caption Loss $\loss_{caption}\left(\bI, \by, \btheta\right)$.}
When training a network to output a reasonable caption, the order of the words is important.
However, in our task, this is not necessarily so, as we only want the contents to be what the caption describes.
We need not regenerate the sentence that the user inputs.
In fact, in our earlier experiments, we found that when the order of the words was considered, depending on the user, the network focused too much on the order of the words, not on the contents, and the performance was poor. 
Thus, for the image caption loss we choose to ignore the order of the words coming out of the captioning network.

Specifically, if we denote the ground truth one-hot encoded vector representation for the $t$-th word of the user caption as $\by_t$, the captioning network as $f\left(\cdot\right)$, and cross-entropy as $H$, we write
\begin{equation}
\vspace{-0.1cm}
    \loss_{caption}\left(\bI, \by, \btheta\right) = 
    H\left(
    \frac{1}{T_{u}}\sum_{t=1}^{T_{u}}{\by_{t}},
    \frac{1}{T_{c}}\sum_{t=1}^{T_{c}}{
        f\left(\bI_{crop}\left(\btheta\right)\right)_{t}
    }
    \right)
    ,
    \label{eq:loss_cap}
\end{equation}
where $T_{u}$ and $T_{c}$ represent the number of words in the user caption and the captioning network $f$ generating caption respectively.
Note that we average the word vectors, effectively removing the order information.

\input{algo/optim/item.tex}

\paragraph{Image Aesthetic Loss \textcolor{red}{ $\loss_{aesthetic}\left(\bI, \btheta\right)$.}}
For the aesthetic term we simply aim to maximize the aesthetic score output from the network. 
If we denote the aesthetic network as $g\left(\cdot\right)$, we therefore write
\begin{equation}
    \vspace{-0.1cm}
    \textcolor{red}{\loss_{aesthetic}\left(\bI, \btheta\right)} = -g\left(\bI_{crop}\left(\btheta\right)\right)
    .
    \label{eq:loss_aes}
\end{equation}
 where, as above, $\bI_{crop}$ is from \eq{crop} and \textbf{y} is the user caption.
This loss helps our crops to be realistic crops close to how humans would crop images as the aesthetic network learned any photographic rules implicitly encoded in professional photographs \cite{ChenACM2017}. 

\input{fig/Heatmap/item.tex}

\subsection{Stabilizing the Optimization}
We design a new cost function considering the two networks' outputs together searching among various bilinear samples.

In Fig. \ref{fig:heatmap} we visualized the total loss space (The extension of this figure can be found in Section 5. of the Supplementary material). During the optimization process, the cropping parameter is shrinking by 2$\%$ in every iteration (See more details in Scale Annealing). The percentages in this figure correspond to the cropping sizes \wrt the original image size. We can see that choosing the correct scale is important in order to ensure that the crop will include all the relevant parts of the image based on the user's description.
However, the image cropping parameter space is too large and non-convex as shown in Fig. \ref{fig:heatmap}, so unstable converge is inevitable.
Our initial attempts indirectly optimizing \eq{objective} were not very successful, even with the help of the scale-space bilinear sampling in \eq{crop}.
We, therefore, propose two additional methods that stabilize the optimization process, leading to better final results.
We explain these methods below and summarize the entire optimization algorithm in \algo{optim}.

\paragraph{Scale Annealing.}
One hardship when directly optimizing for the crop parameters, is that once determined to be outside of the crop region, pixels outside the crop region have no means to affect the optimization process.
This leads to instability when optimizing also for scale, as for example, when the crop accidentally shrinks, it will be difficult for the system to recover from it.
We, therefore, choose to exclude the scale parameter from optimization and to anneal the scale to become smaller throughout the optimization process.
Specifically, we set the scale to be $s^i$, where $i$ is the optimization iteration.
We empirically set $s=0.98$.
In other words, the scale is reduced by 2\% at each iteration.
During optimization, we keep tracking which crop parameter gave the lowest loss so far and return that crop region as our final result.

\paragraph{Optimizing with Multiple Restarts.} 
To further stabilize the optimization process and escape local optima, we employ a multiple restart technique~\cite{multistart} as shown in Fig. \ref{fig:multiple_restarts}.
Based on the research of \cite{kwedlo2015new}, for each optimization iteration, we apply random noise $\Delta{x}\sim\calU\left(0,1\right)$ and $\Delta{y}\sim\calU\left(0,1\right)$, where $\calU\left(0,1\right)$ is the uniform distribution,
to the outcome of the previous iteration. 
After adding the noise, we further clip them to prevent the cropped region from going out of the image border.
We then run our optimization to find the optimal crop for this given scale, and repeat the process $K$ times, and average their results to obtain our final solution for this scale (in our experiments, we use the $K = 10$).

\input{fig/multiple_restart/item.tex}

If we denote the crop center estimates at iteration $i$ as $x^i$ and $y^i$, we write
\vspace{-0.1cm} 
\begin{equation}
    \left(x^{i+1}, y^{i+1}\right) = \frac{1}{K}\sum_{k=1}^K
    \left(\hat{x}_k^i, \hat{y}_k^i\right)
    ,
    \label{eq:restart_collect}
    \vspace{-0.4cm}
\end{equation}
where
\begin{equation}
\vspace{-0.1cm}
    (\hat{x}_k^{i}, \hat{y}_k^{i}) = 
    \argmin_{x, y|x_0\sim\calU\left(0,1\right), y_0\sim\calU\left(0,1\right)}
    \loss_{total}\left(\bI, \by, \left(x, y, s^i\right)\right)
    ,
    \label{eq:restart}
\end{equation}
and $x_0$ and $y_0$ are the starting points of optimization for $x$ and $y$ respectively, and $x_i$ and $y_i$ are the final converged locations for this scale level.
To find the \textbf{$\hat{\theta}$}, we apply limited-memory Broyden-Fletcher-Goldfarb-Shanno (L-BFGS) \cite{byrd1995limited}, as we are searching for the optimal location and not simply seeking to perform gradient descent for each scale. The L-BFGS approach is one of the most popular quasi-Newton methods that construct positive definite Hessian approximations. It is a local search algorithm that is intended for convex optimization problems with a single optimum. Thus, the L-BFGS is suitable for our method, as we apply scale annealing and prefer local optimizations.

Although the multiple restart strategy does not theoretically guarantee that the average location is lower in terms of the final objective, it ensures that we are always optimizing towards the general improving direction.

%% file: fig/framework/item.tex
\begin{figure*}[t]
\centering
\includegraphics[width=\linewidth]{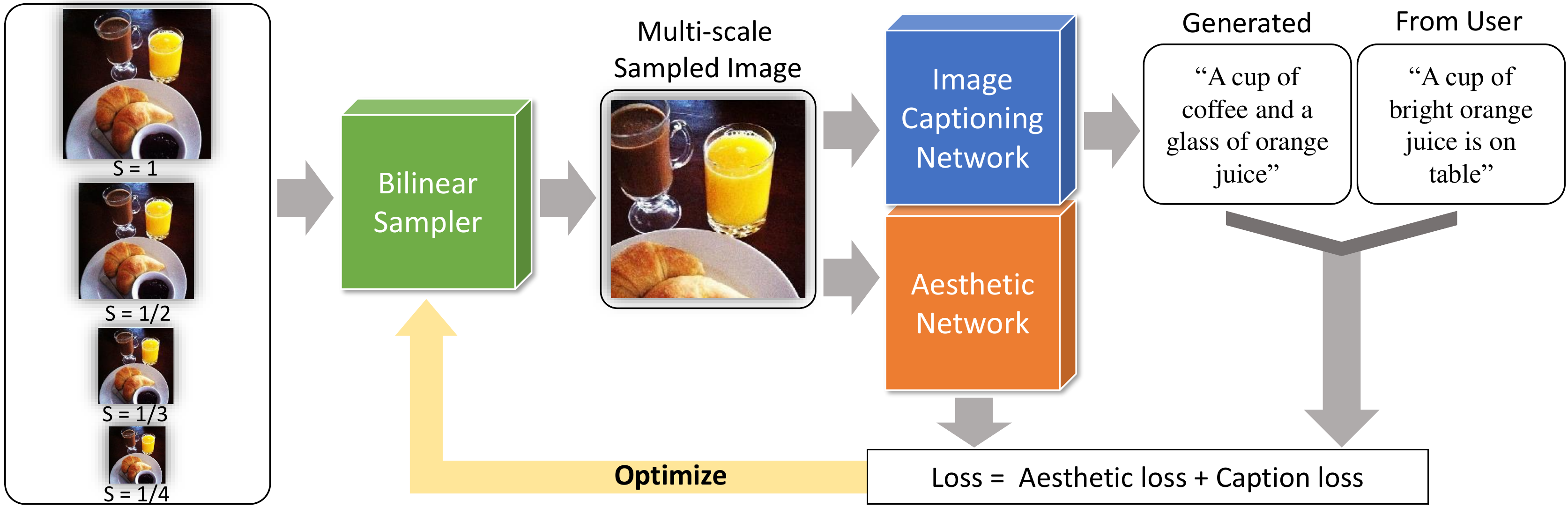}
\caption{Overall framework of the proposed method (CAGIC). The framework takes an image as input, which goes through multi-scale bilinear sampling to produce a cropped image.
Note that the parameters for this sampling do not come from a network as in other existing works, but rather are the parameters that we will directly optimize for later.
We then use this cropped region as input to both the image captioning network and the aesthetic network.}
\vspace{-0.3cm}
\label{fig:framework}
\end{figure*} 

%% file: algo/optim/item.tex
  \begin{algorithm}[b!]
  \caption{Optimization with multiple restart.}
  \label{alg:optim}
  \begin{algorithmic}[1]
    \Require{
      $\bI$ : input image,
      $\by$ : user caption
    }
    \Function{Optimize}{$\bI$, $\by$}
    \For{$i=1$ to $S$}
    \Comment{For each scale level}
    \For{$k=1$ to $K$}
    \Comment{Optimize $K$ times}
        \State{$x_0 \sim \calN\left(x^i, \sigma\right)$}
        \Comment{Restart initialization}
        \State{$y_0 \sim \calN\left(y^i, \sigma\right)$}
        \State{$\hat{x}_k^{i}, \hat{y}_k^{i} \leftarrow$ \eq{restart}}
        \Comment{Find optimum point}
    \EndFor
    \State{$x^{i+1} \leftarrow \frac{1}{K}\sum_{k=1}^K \hat{x}_k^i$}
    \Comment{Gather result}
    \State{$y^{i+1} \leftarrow \frac{1}{K}\sum_{k=1}^K \hat{y}_k^i$}
    \EndFor

     \EndFunction

  \end{algorithmic}
\end{algorithm}

%% file: fig/Heatmap/item.tex
\begin{figure*}
\centering
\includegraphics[width=\linewidth]{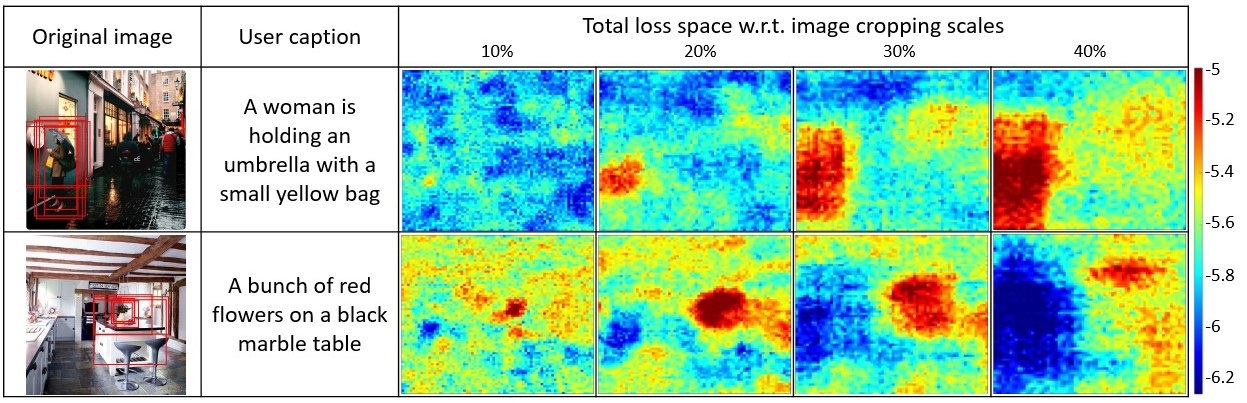}
\caption{
We visualize the total loss space for different image cropping parameters. The percentages indicate the image cropping scales compared to the original image size.}

\vspace{-0.4cm}
\label{fig:heatmap}
\end{figure*}

%% file: fig/multiple_restart/item.tex
\begin{figure}
\centering
\includegraphics[width=0.85\linewidth]{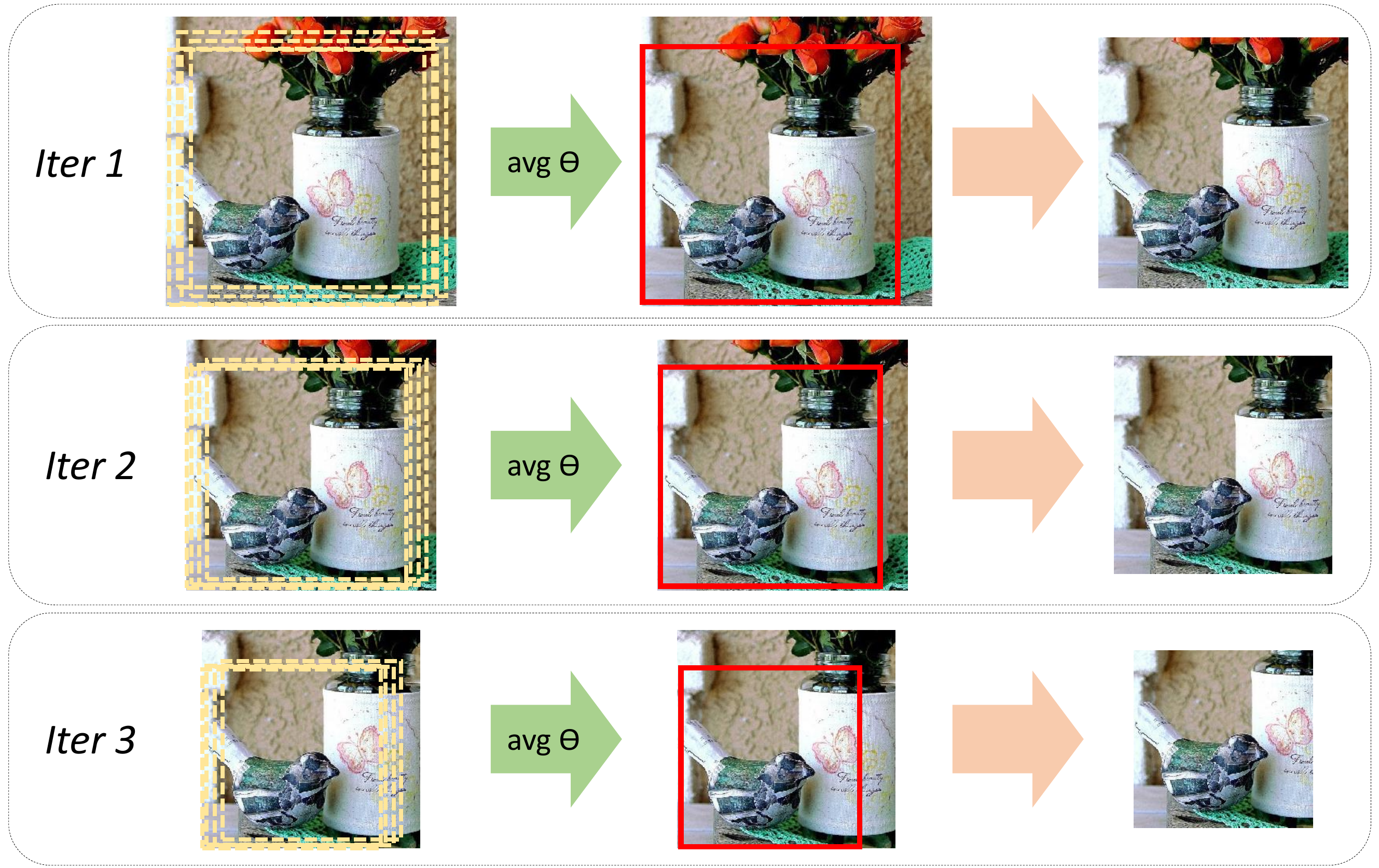}
\caption{
For each iteration, we employ the multiple restart strategy which starts random location and takes the average of the optimization outcomes. In the next iteration, the scale is annealed as scheduled.
}
\vspace{-0.3cm}
\label{fig:multiple_restarts}
\end{figure}

%% file: tex/result.tex
\section{Experiments}

We implement our method
in TensorFlow~\cite{Tensorflow}\footnote{To guarantee reproducibility, we will release our code if the paper is accepted}.
All experiments are run on
an Intel i7- CPU @ 3.40GHZ, 16 GB RAM, and two NVIDIA TITAN Xp GPU.

\subsection{Dataset and baselines}

As the task we aim for -- cropping based on natural descriptions -- is different from what existing datasets can offer, we create a novel dataset.
Before we discuss our dataset in detail, we first briefly review existing datasets.

\paragraph{Deficiencies of existing datasets.}
A widely used dataset for natural images with captions is the MS-COCO dataset \cite{Lin2014MicrosoftCC}, which  contains 123k images, with predefined splits for training, validation, and testing. 
Each image is annotated with five captions by Amazon Mechanical Turkers.
However, most of the captions describe the whole image, and is not suitable for caption-guided cropping.
Therefore, based on MS-COCO, RefCOCO~\cite{kazemzadeh2014referitgame} dataset was proposed, where parts of images are described with annotated bounding boxes.
This dataset, however, is heavily skewed towards tight detection of certain objects, and are not suitable for creating natural crops. 
They are aimed more at text-guided object detection.

\input{tbl/dataset/item.tex}

In \fig{dataset} we show example captions on the same image from different datasets to demonstrate that captions from refCOCO~\cite{kazemzadeh2014referitgame}, MS-COCO~\cite{Lin2014MicrosoftCC} are not adequate for our tasks.
As one can see, the caption from MS-COCO describes the entire image, while the captions
from refCOCO is very short and related to a single object only.

\begin{figure}[b]
    \centering
    \includegraphics[width = 0.32\textwidth]{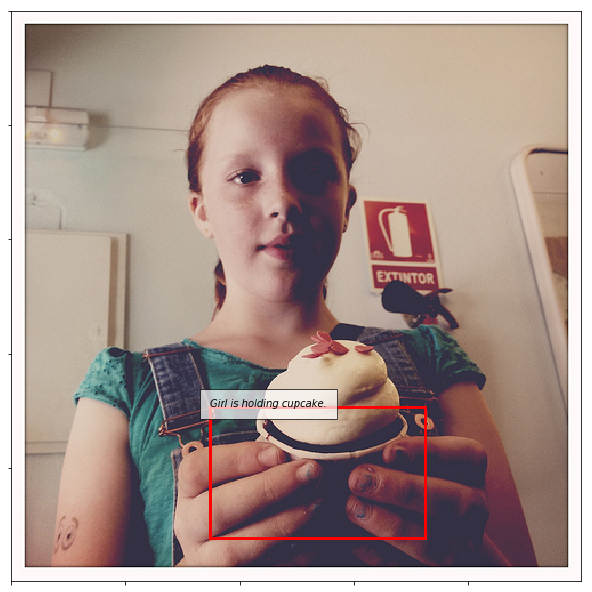}
    \includegraphics[width = 0.32\textwidth]{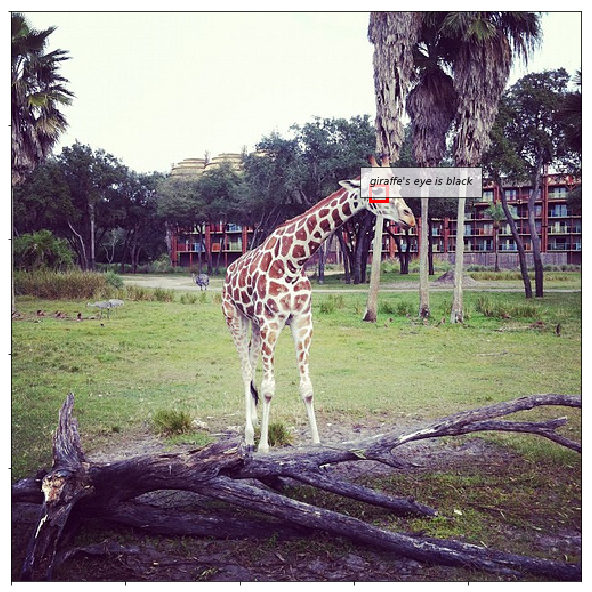}
    \includegraphics[width = 0.32\textwidth]{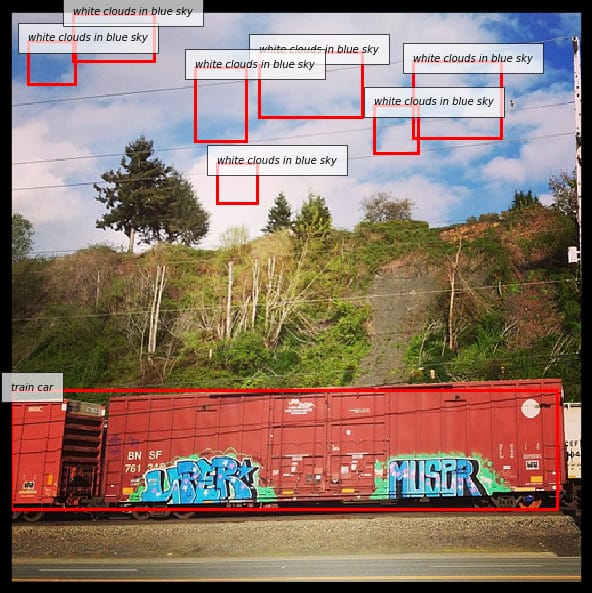}
    \caption{Example captions and corresponding annotations from the Visual Genome dataset.}
    \label{fig:vg}
\end{figure}

The Visual Genome dataset~\cite{krishnavisualgenome} is perhaps the closest to what we need for our task.
This large dataset consists of 108k images from the YFCC100M~\cite{thomee2015yfcc100m} and MS-COCO datasets.
Different from the original captions of the MS-COCO dataset, the Visual Genome provides about 50 descriptions of different image regions. The annotations for this dataset is also geared towards more traditional localization and description tasks.
However, there are two problems in this dataset that makes it not feasible for evaluating natural crops: (i) annotations are tight bounding boxes focused on objects -- that do not represent how people typically take photos; (ii) identical captions can denote multiple regions -- for example ``red sky'' could correspond to any bounding box within the sky -- demonstrating that annotations are only part of the possible ``ground truth'', and not overlapping with any of these do not mean that a crop is wrong.

As shown in \fig{vg}, the captions and annotations would not look at all like natural crops, if they were to be cut out -- they are either too tight and sometimes not even capturing the entire object, as shown by the examples. 
Also notice that the captions are very short, almost as if they are descriptions of a single object.
There is also an issue of non-specific regions as in the case of \fig{vg} right, where the descriptions all talk about the sky and the cloud -- in addition, none of the crops would actually be recognizable to a human being on what they are about. In the supplementary appendix, we show qualitative and quantitative results on this dataset.

\input{fig/GTbbox/item.tex}

\subsubsection{A novel dataset}
Therefore, we create a novel dataset based on MS-COCO, similarly to the refCOCO and Visual Genome datasets.
We select 100 images randomly from the MS-COCO test set to avoid the images being ever seen by any of our pre-trained networks and create our own captions for each image.

\paragraph{Ground-truth captions.}
In our new dataset, we are interested in evaluating the ability to automatically generate crops related to a given caption.
We therefore manually generate descriptive expressions, focusing on distinctive parts of the image.
Our captions are roughly 10 words long on average to ensure that they are specific and clear.

\paragraph{Ground-truth crops.}
One of the tricky parts in creating a dataset for caption-based image cropping is the definition of ground truth.
The concept of ``which image region fits the description well'' is subjective and can wildly differ from person to person.
Therefore, to remove the subjective nature of our ground truth annotations as much as possible, 
we carefully select regions of the image that are unique and distinctive for a human annotator to create a ``natural'' crop out of.
We then
asked 8 participants to generate ground truth crops based on our captions individually.
We further asked the participants to consider the aesthetics of these crops. 
We use these ground truth annotations to perform quantitative comparisons.
As an evaluation metric we use the Intersection over Union (IoU), a standard metric for evaluating bounding box-based tasks~\cite{ChenACM2017, overlap}.

Even with care, it is inevitable that the cropping task is subjective and dependent on the annotator. 
We show an example of the ground-truth annotations in \fig{overlap_gt} (top), as well as the agreement between annotators in \fig{overlap_gt} (bottom). 
Note that even when disagreeing on the exact crop, they are all overlapping -- the main content is shared.

\subsubsection{Baselines}
We compare our method against the following eight different methods: 
{\bf GradCAM~\cite{Selvaraju2017GradCAMVE}} -- a naive untrained baseline where we apply GradCAM with the captioning network to extract regions in the image corresponding to the user caption. We then threshold the activation map with a threshold of 0.2 of the maximum value, which we empirically set;
{\bf A2-RL~\cite{li2018a2}, VPN~\cite{goodview2018}, Anchor~\cite{Zeng_2019_CVPR}} -- full automatic methods to demonstrate that these crops do not necessarily correspond to the users' intentions;
{\bf GradCAM+A2-RL/VPN/Anchor} -- GradCAM and the auto-crop networks combine sequentially;
{\bf MAttNet~\cite{mattnet}} -- the state-of-the-art network for tight localization from text descriptions.

\subsection{Comparison with the state-of-the-art}

\input{fig/same_img/item.tex}

\input{tbl/qualitative/item.tex}

\paragraph{Qualitative results.}
We first show qualitative highlights in \fig{qualitative}. As shown, our method provides crops of the highest quality.
In \fig{same_img}, we show that our method can deliver entirely different results on the same image, when different captions are provided.
Notice how our results are not only corresponding well to the provided captions, but also good looking, demonstrating that our method successfully considers both aspects.
In particular, as shown in \fig{same_img} (b) and (f), our method well crops regions when the description is about parts of the image that are small or are located near the edge.

\paragraph{Quantitative results -- IoU.}
We report the average IoU between produced crops and all eight of the ground-truth annotations in \tbl{overlap}. As shown, our method provides the highest IoU value among the compared methods. 
Note that we have higher numbers than even those which were trained specifically for referencing from captions.
This is mainly due to the fact that existing methods perform tight crops. However, when a human is asked to perform the same task, they tend to include context.

\input{tbl/quantitative/item.tex}

\input{tbl/quantitative/item2.tex}

\paragraph{User study -- is the crop really what we want?}
As one of the main goals here is to faithfully follow user guidance, we evaluate it by asking users to re-annotate the crop outcomes.
We asked four users, who were not exposed previously to the original images, to caption cropped images with natural descriptions.
We then evaluate how similar these new descriptions are to the target descriptions of our dataset.
We evaluate only the top three methods from the IoU evaluation.
We report results in terms of metrics used for natural language processing\footnote{\scriptsize{\url{https://github.com/Maluuba/nlg-eval}}} in \tbl{crosscrop}.
Our method provides the best results for all metrics, demonstrating that the user intention is best preserved.

\input{tbl/user/item.tex}
\paragraph{User study -- which crop is better?}

Due to the subjective nature of our task, we further perform a user study, where we ask users to select which image is preferred over the top-3 methods, as well as the original image as the baseline.
Specifically, we ask users to \textit{``Select the crop described by the caption which looks the best,''} given the four images in a graphical user interface.
We ask a total of 13 users, resulting in more than 1000 decisions.

In the user study, for each user, we show a randomly selected subset of our dataset. 
We report the probability of being selected for each method in \tbl{user}.
Our method is the most preferred among compared methods.
Our user study shows that the subjects preferred the cropped images over the original image. Users preferred the methods with wider output crops that were not mainly focused on the subject of the image caption; therefore, they were able to preserve the contextual information. While the output of GradCAM was contextually correct, users tend to select the proposed method when the aesthetics of the crop was not pleasing. This highlights that aesthetics need to be considered when emulating what humans would do.

\input{fig/ablation/item.tex}
\input{fig/Aesthetics_ablation/item.tex}

\subsection{Ablation Study}
To motivate our design choices, we present a qualitative comparison of our method with various components of our full pipeline disabled; see~\fig{ablation}.
We compare against five variants. The full method uses the caption network (CN) along with scale anneal (SA), multiple restart (MR), multi-scale bilinear sampling (MS) as well as aesthetics.
As shown,
as we introduce aesthetics scores to the system we are able to produce more visually appealing crops of the original image. 
In the case of CN+SA+MR+MS, the method is able to find the relevant part of the original image, but the contents are not placed in the centre, as can be noticed in the examples shown in the last two rows.
Multiple restart help the approach to centre the content better, while scale annealing prevents drastic scale changes.
In short, the full method delivers the best outcome in all cases.

The quantitative results of the ablation study are in Table \ref{tbl:ablationquantitative}. show that every component of the proposed method is effective and that using aesthetics on top of the different optimization and search stabilization methods improved the results.
In addition, we further demonstrate the effectiveness of the aesthetics loss in our framework in \fig{aesth_ablation}. 
As shown, the produced crops are most similar to the ground truth regions when the aesthetics loss is enabled.

\input{tbl/ablation/item_quantitative}

\subsection{Runtime}

Currently, our un-optimized implementation is unable to run real-time –- requires an average of 2.06 seconds per single optimization iteration. The runtime of the other deep learning-based methods where the input goes through one pre-trained network is shorter compared to our optimization-based method. However, despite our method not being real-time, it was clearly demonstrated that it performs better and it is flexible.
Our future work would be to resolve this, similar to how style transfer \cite{Gatys16a} started off taking minutes per image, but is now able to run real-time \cite{Johnson16}.

%% file: tbl/dataset/item.tex
 \begin{table}[!h]
 \begin{center}

\includegraphics[width=0.7\linewidth]{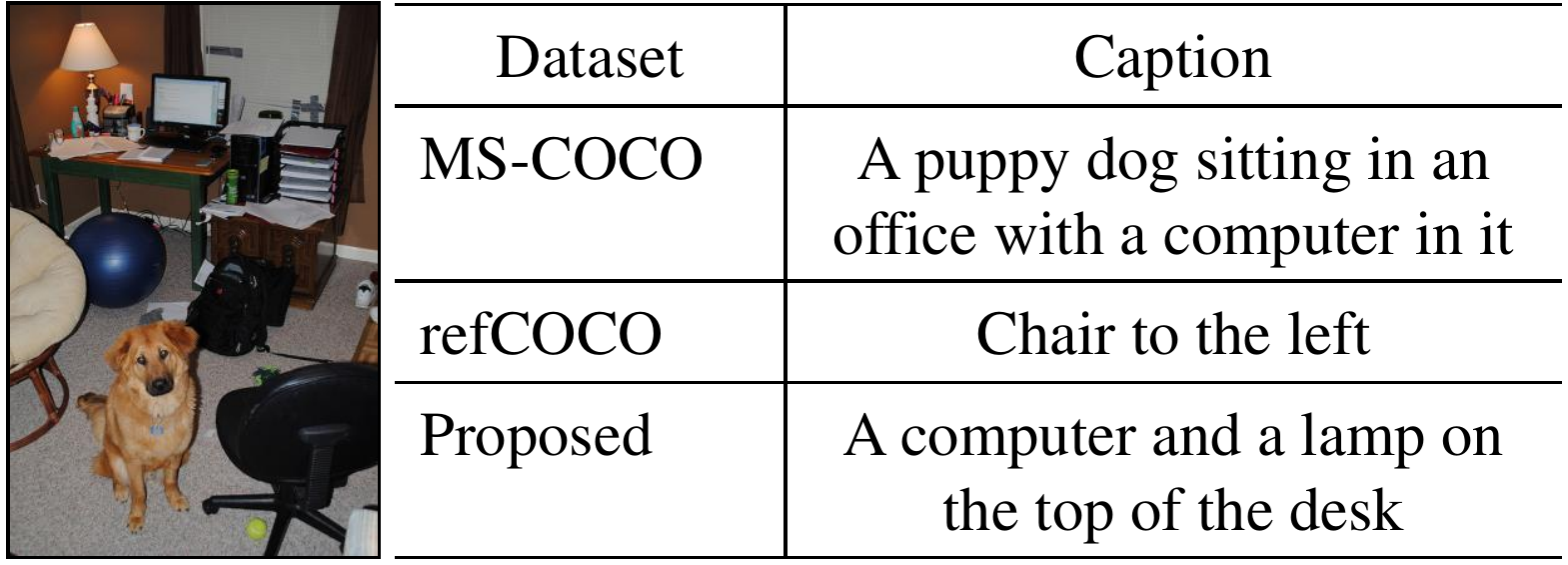}
\captionof{figure}{Caption comparison between MS-COCO, refCOCO, and our new dataset on ImgId=38353 of the MS-COCO train2014 set.}
\label{fig:dataset}
 \end{center}
 \end{table}

%% file: fig/GTbbox/item.tex
\begin{figure}
\centering
\includegraphics[width=0.8\linewidth]{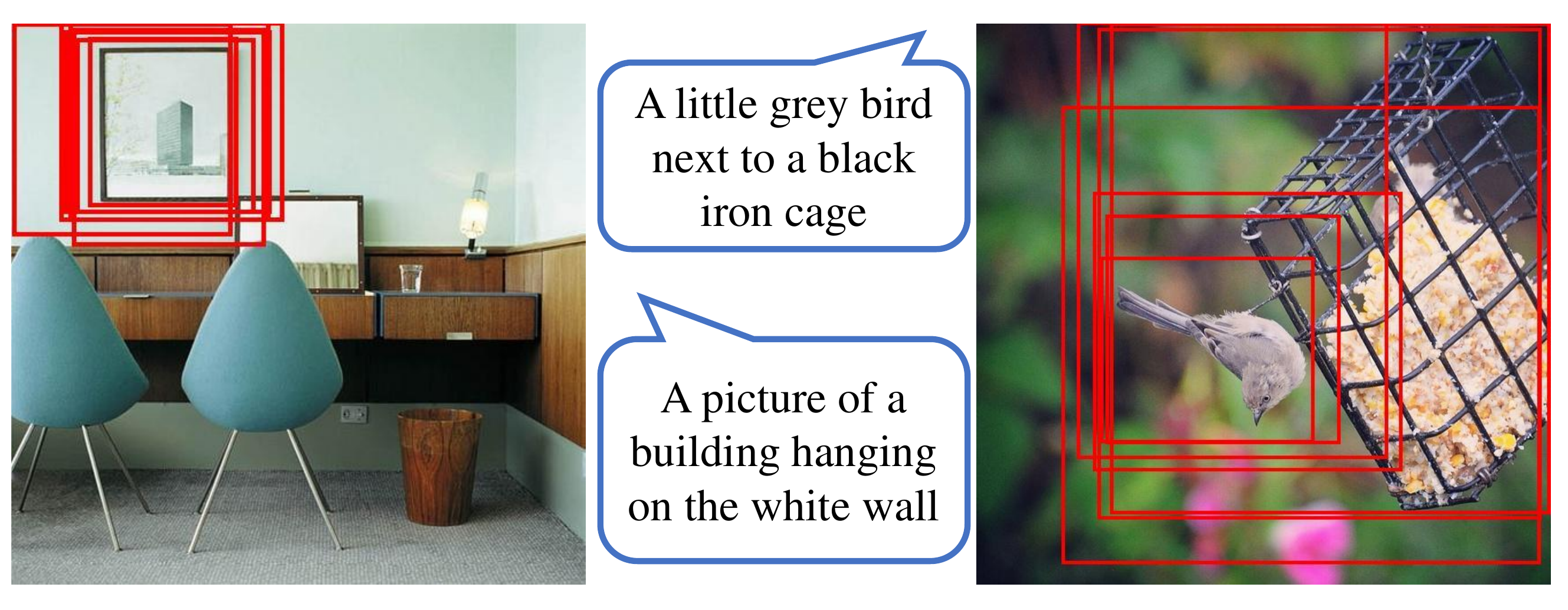}
\includegraphics[width=0.85\linewidth]{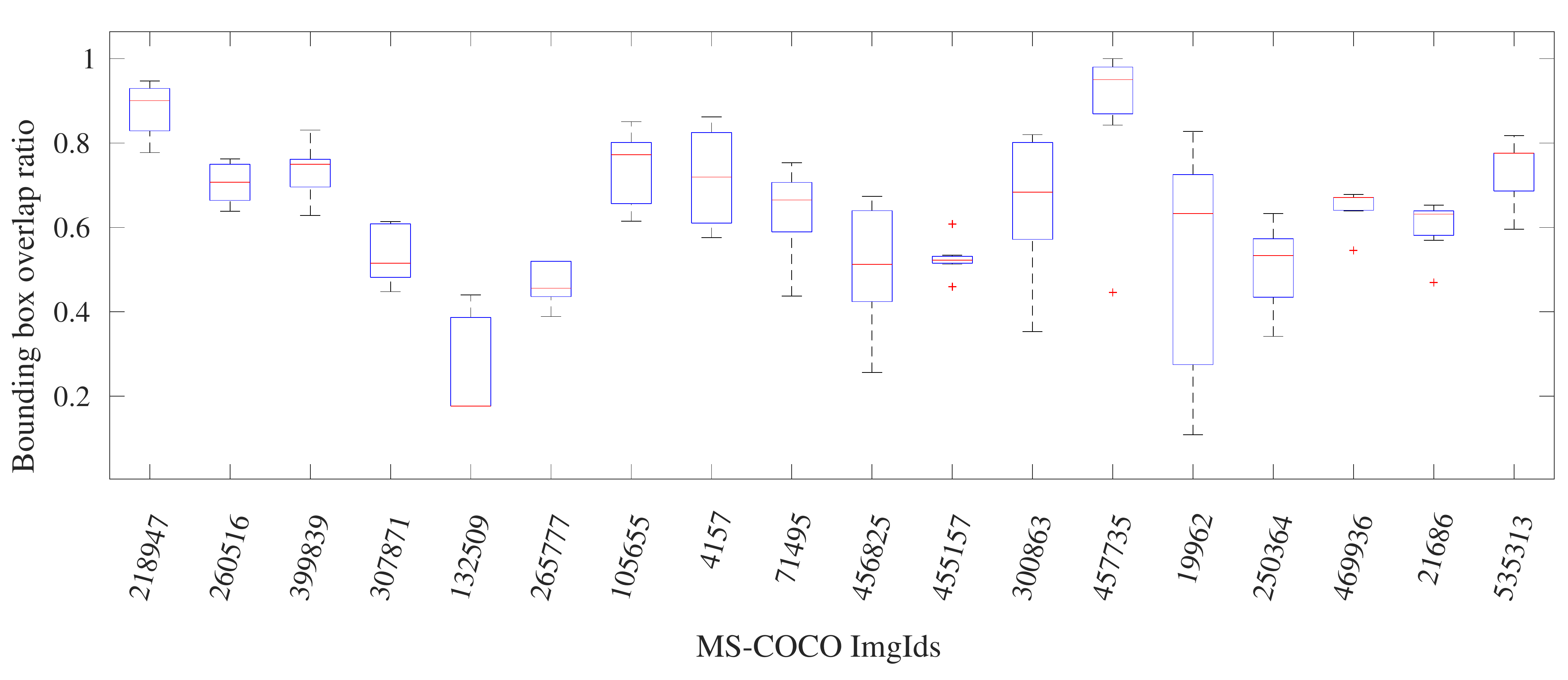}
\caption{(Top) Illustration of the diversity of the bounding box annotations for different images. (Bottom) Box plots of the distribution of ground truth overlaps for different MS-COCO images.}
\label{fig:overlap_gt}
\end{figure}

%% file: fig/same_img/item.tex
\def \figsame {0.30}
\begin{figure}[t]
\centering
\begin{subfigure}[t]{\figsame\linewidth}
    \frame{\includegraphics[width=0.95\linewidth, trim = -0 0 -0 0, clip]{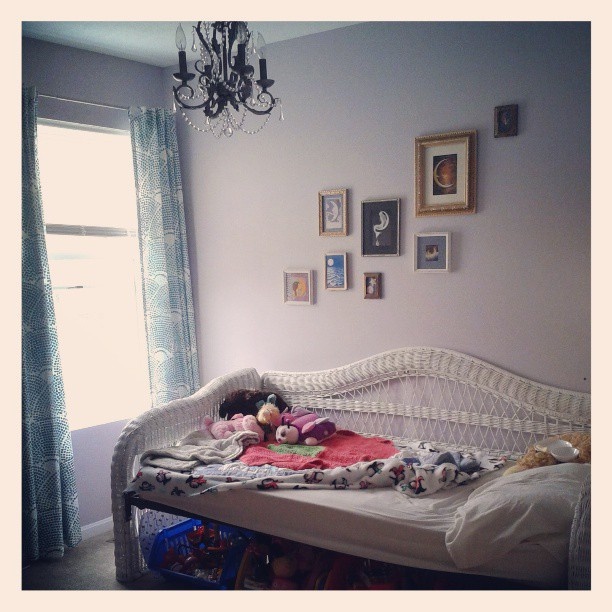}}
    \vspace{-1mm}
    \caption{\scriptsize{Original image}}
\end{subfigure}
\hspace{1mm}
\begin{subfigure}[t]{\figsame\linewidth}
    \frame{\includegraphics[width=0.95\linewidth, trim = -0 0 -0 0, clip]{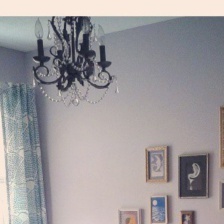}}
    \vspace{-1mm}
    \caption{\scriptsize{A ceiling lamp on the top of the room}}
\end{subfigure}
\hspace{1mm}
\begin{subfigure}[t]{\figsame\linewidth}
    \frame{\includegraphics[width=0.95\linewidth, trim = -0 0 -0 0, clip]{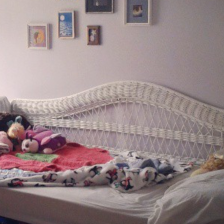}}
    \vspace{-1mm}
    \caption{\scriptsize{Some dolls and a red blanket are on the bed}}
\end{subfigure}

\begin{subfigure}[t]{\figsame\linewidth}
    \frame{\includegraphics[width=0.95\linewidth, trim = -0 0 -0 0, clip]{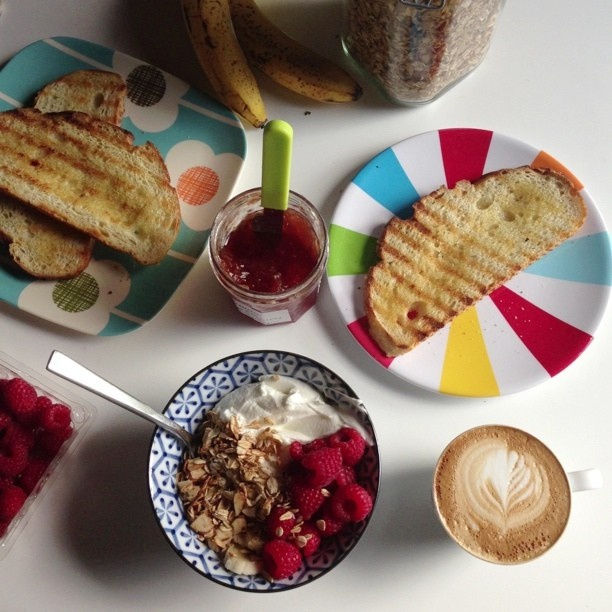}}
    \vspace{-1mm}
    \caption{\scriptsize{Original image}}
\end{subfigure}
\hspace{1mm}
\begin{subfigure}[t]{\figsame\linewidth}
    \frame{\includegraphics[width=0.95\linewidth, trim = -0 0 -0 0, clip]{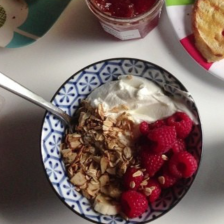}}
    \vspace{-1mm}
    \caption{\scriptsize{A blue bowl of white cream and red berries with a metal spoon in it}}
\end{subfigure}
\hspace{1mm}
\begin{subfigure}[t]{\figsame\linewidth}
    \frame{\includegraphics[width=0.95\linewidth, trim = -0 0 -0 0, clip]{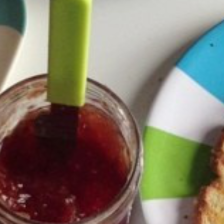}}
    \vspace{-1mm}
    \caption{\scriptsize{A glass of red jam with a green spoon}}
\end{subfigure}
\caption{Different captions can lead to entirely different crops even on a single image.}
\label{fig:same_img}
\end{figure}

%% file: tbl/qualitative/item.tex
\begin{figure*}
\centering
    \setlength\tabcolsep{1.3pt}
\resizebox{1.0\linewidth}{!}{
     \begin{tabular}{P{1.5cm}P{2.0cm}|P{1.5cm}P{1.5cm}P{1.5cm}P{1.5cm}P{1.5cm}P{1.5cm}P{1.5cm}P{1.5cm}P{1.5cm}}\toprule
       Original Image & User Caption & \mbox{A2-RL}\cite{li2018a2} & VPN \cite{goodview2018}  & Anchor \cite{Zeng_2019_CVPR}&  GradCAM +A2-RL& GradCAM +Anchor & GradCAM +VPN & GradCAM \cite{Selvaraju2017GradCAMVE} & MAttNet \cite{mattnet} & CAGIC \\\midrule

       \includegraphics[width=\linewidth,height=1.5cm,keepaspectratio]{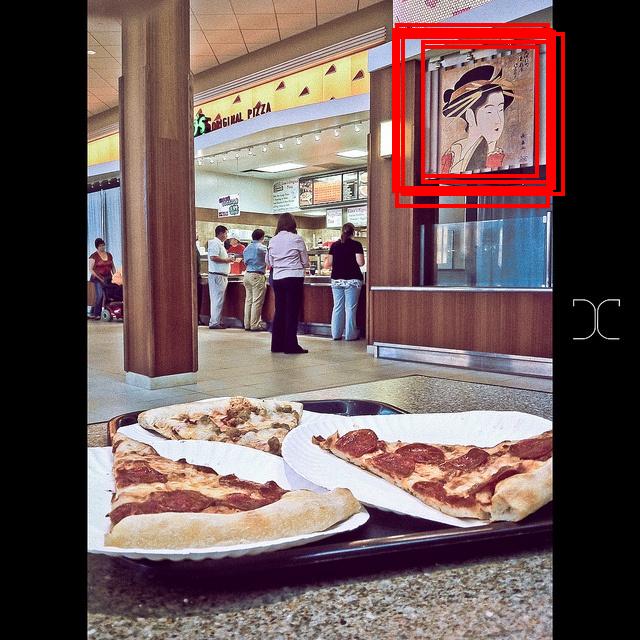} & 
       \vspace{-1.5cm}\scriptsize A japanese style painting of woman is on the wall  & 
       \includegraphics[width=\linewidth,height=1.5cm,keepaspectratio]{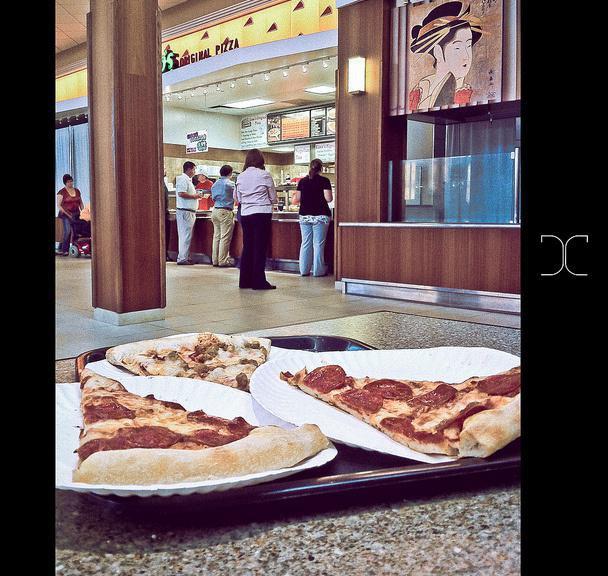} & \includegraphics[width=\linewidth,height=1.5cm,keepaspectratio]{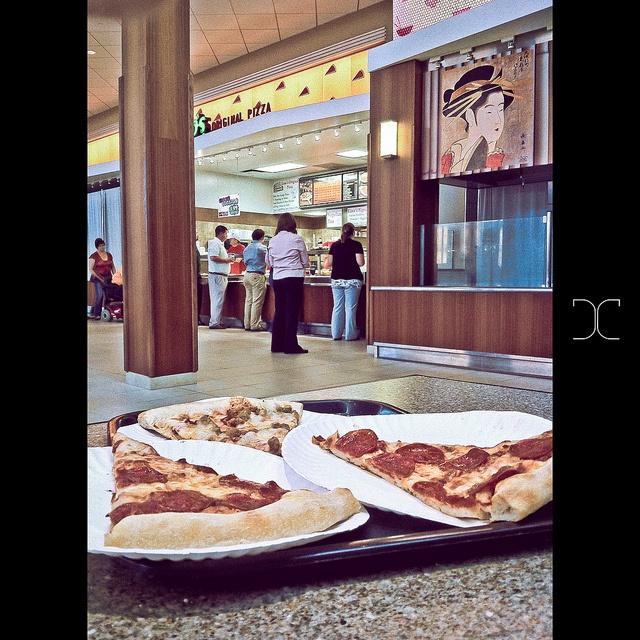} & \includegraphics[width=\linewidth,height=1.5cm,keepaspectratio]{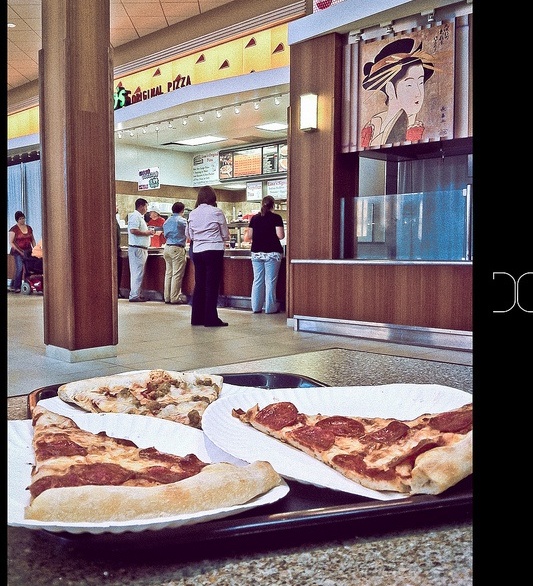} & \includegraphics[width=\linewidth,height=1.5cm,keepaspectratio]{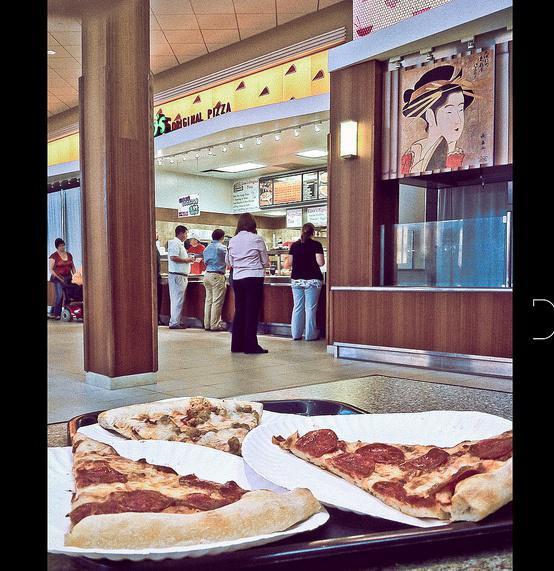} & \includegraphics[width=\linewidth,height=1.5cm,keepaspectratio]{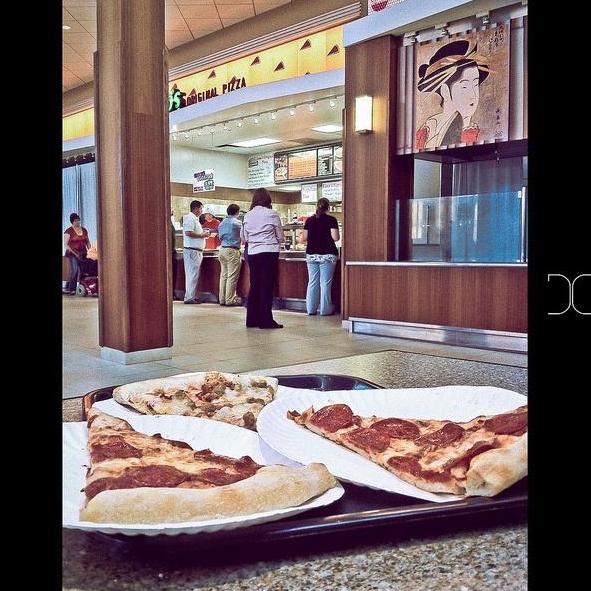} & \includegraphics[width=\linewidth,height=1.5cm,keepaspectratio]{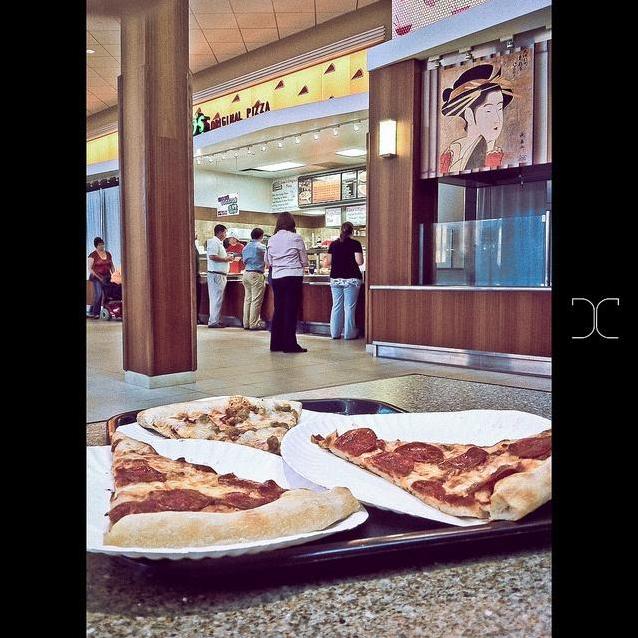} & \includegraphics[width=\linewidth,height=1.5cm,keepaspectratio]{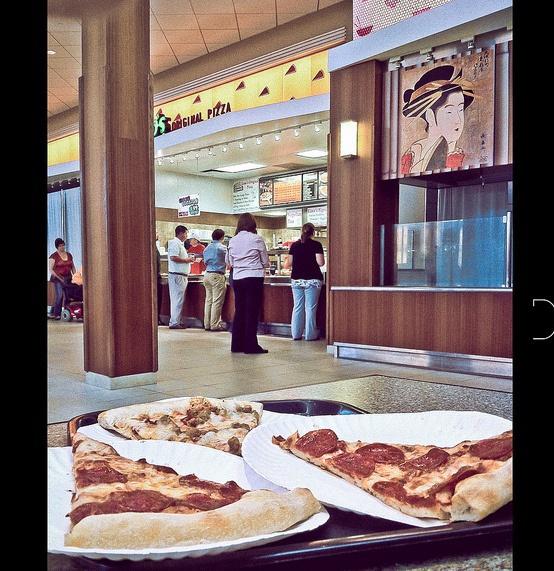} & \includegraphics[width=\linewidth,height=1.5cm,keepaspectratio]{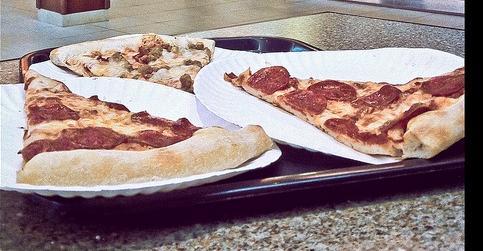} & \includegraphics[width=\linewidth,height=1.5cm,keepaspectratio]{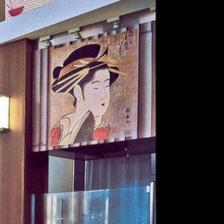} \\

       \includegraphics[width=\linewidth,height=1.5cm,keepaspectratio]{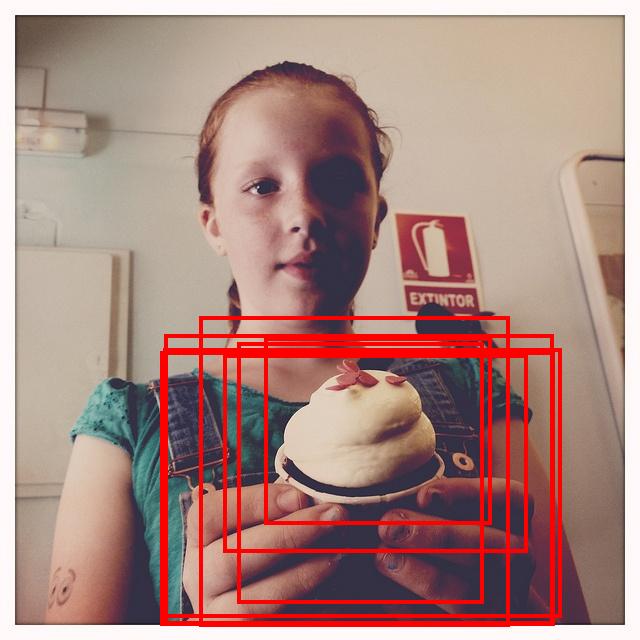} &
       \vspace{-1.4cm}\scriptsize Hands holding a white cupcake & 
       \includegraphics[width=\linewidth,height=1.5cm,keepaspectratio]{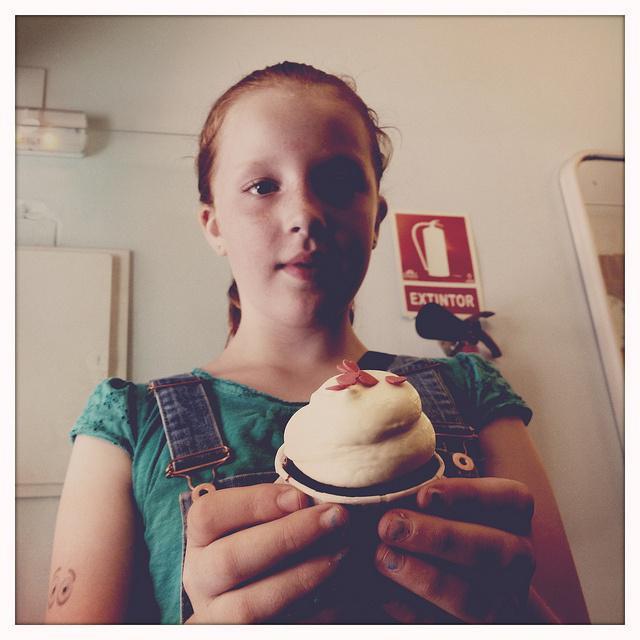} & \includegraphics[width=\linewidth,height=1.5cm,keepaspectratio]{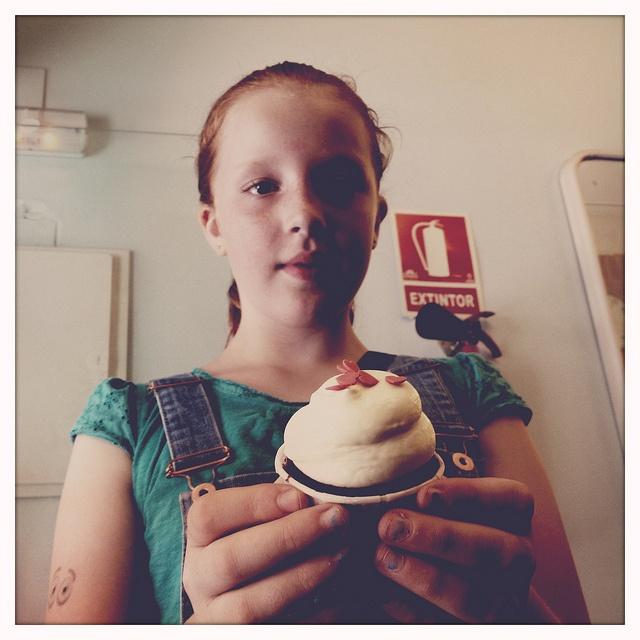} & \includegraphics[width=\linewidth,height=1.5cm,keepaspectratio]{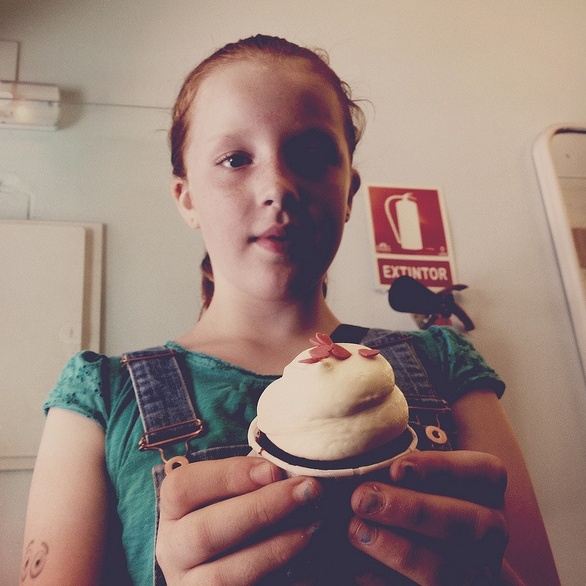} & \includegraphics[width=\linewidth,height=1.5cm,keepaspectratio]{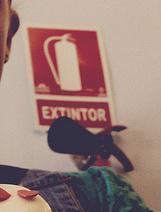} & \includegraphics[width=\linewidth,height=1.5cm,keepaspectratio]{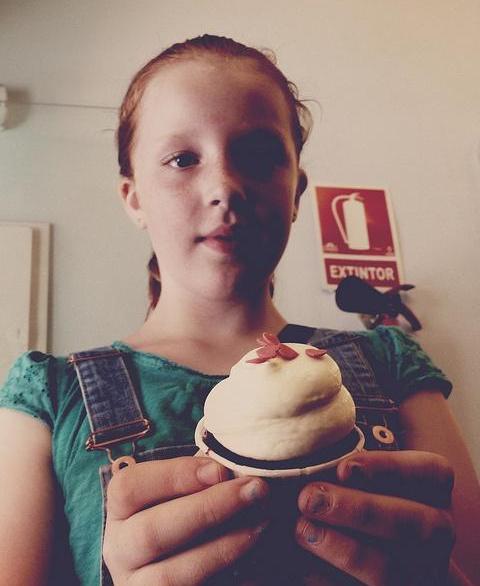} & \includegraphics[width=\linewidth,height=1.5cm,keepaspectratio]{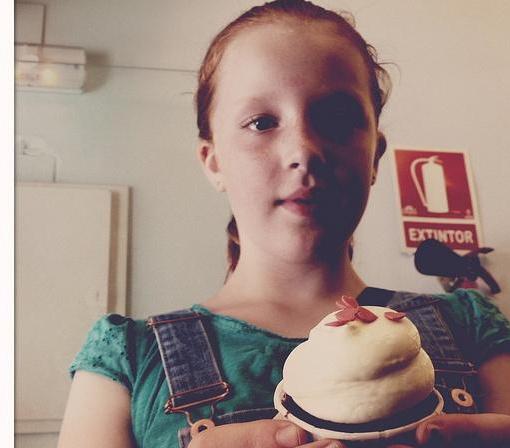} & \includegraphics[width=\linewidth,height=1.5cm,keepaspectratio]{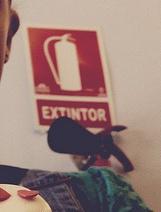} & \includegraphics[width=\linewidth,height=1.5cm,keepaspectratio]{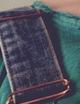} & \includegraphics[width=\linewidth,height=1.5cm,keepaspectratio]{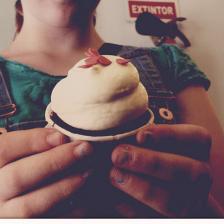}\\

       \includegraphics[width=\linewidth,height=1.5cm,keepaspectratio]{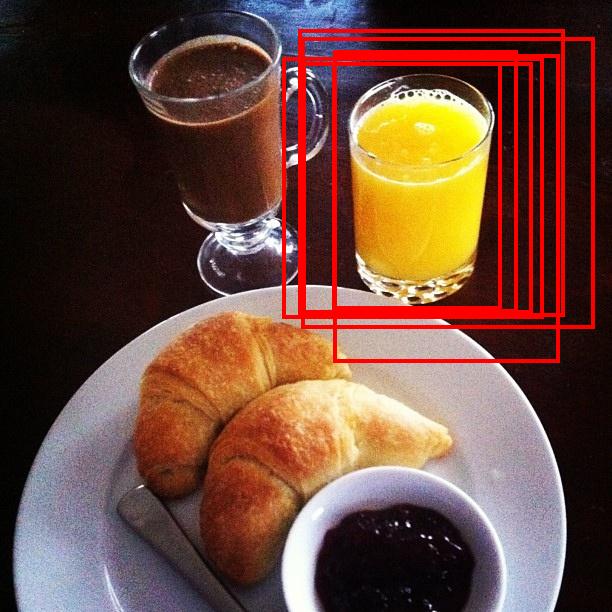} & 
       \vspace{-1.5cm}\scriptsize A cup of bright orange juice is on table & 
       \includegraphics[width=\linewidth,height=1.5cm,keepaspectratio]{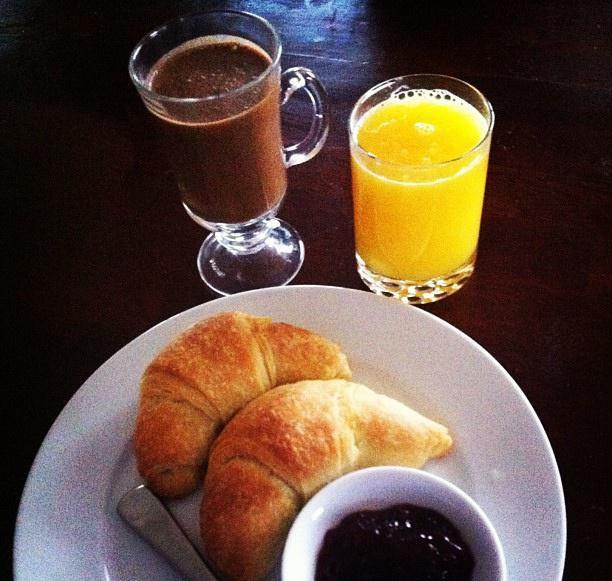} & \includegraphics[width=\linewidth,height=1.5cm,keepaspectratio]{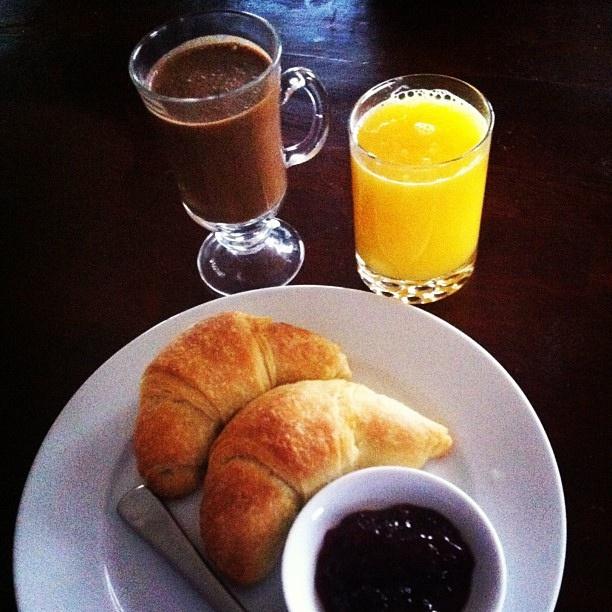} & \includegraphics[width=\linewidth,height=1.5cm,keepaspectratio]{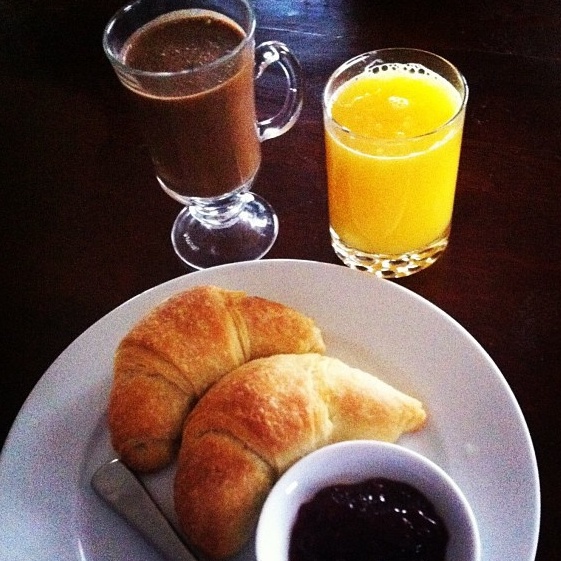} & \includegraphics[width=\linewidth,height=1.5cm,keepaspectratio]{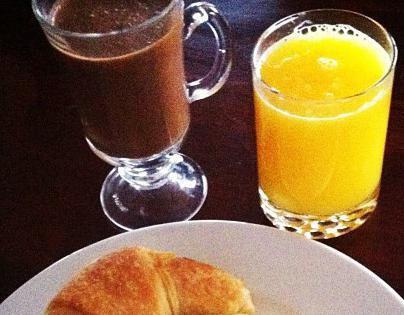} & \includegraphics[width=\linewidth,height=1.5cm,keepaspectratio]{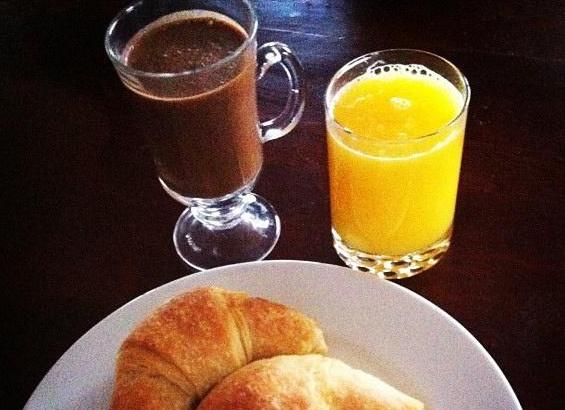} & \includegraphics[width=\linewidth,height=1.5cm,keepaspectratio]{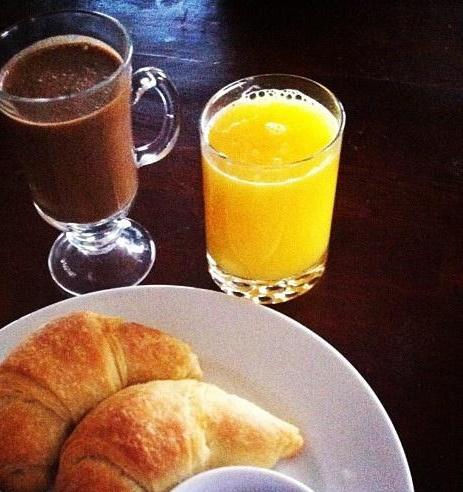} & \includegraphics[width=\linewidth,height=1.5cm,keepaspectratio]{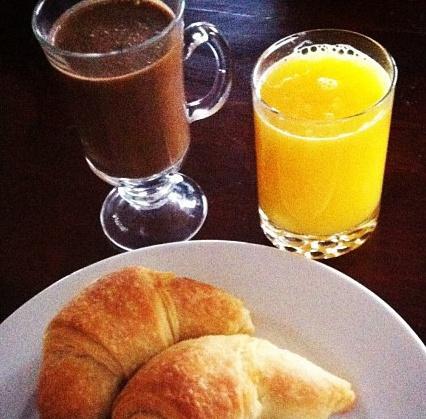} & \includegraphics[width=\linewidth,height=1.5cm,keepaspectratio]{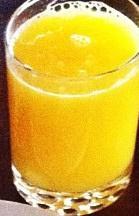} & \includegraphics[width=\linewidth,height=1.5cm,keepaspectratio]{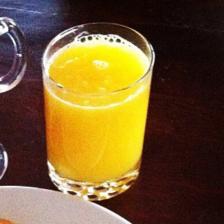}\\

      \includegraphics[width=\linewidth,height=1.5cm,keepaspectratio]{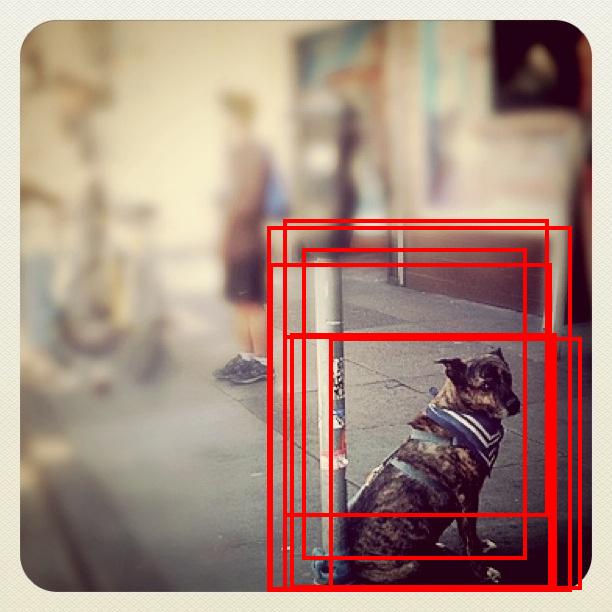} & 
      \vspace{-1.5cm}\scriptsize A black dog is wearing a tie beside a pole & 
      \includegraphics[width=\linewidth,height=1.5cm,keepaspectratio]{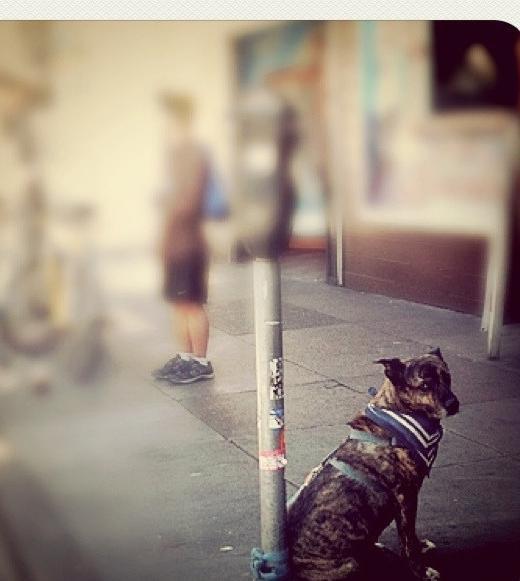} & \includegraphics[width=\linewidth,height=1.5cm,keepaspectratio]{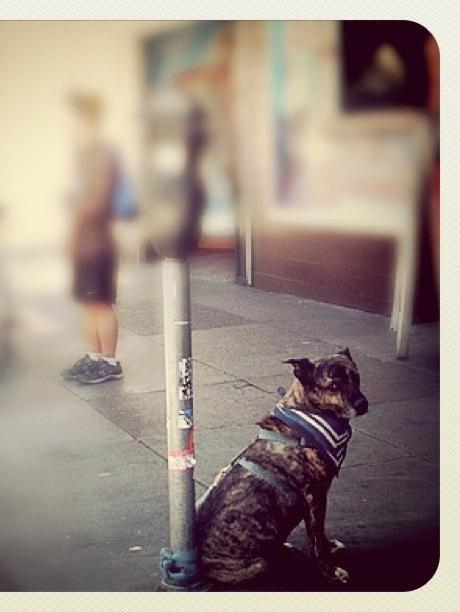} & \includegraphics[width=\linewidth,height=1.5cm,keepaspectratio]{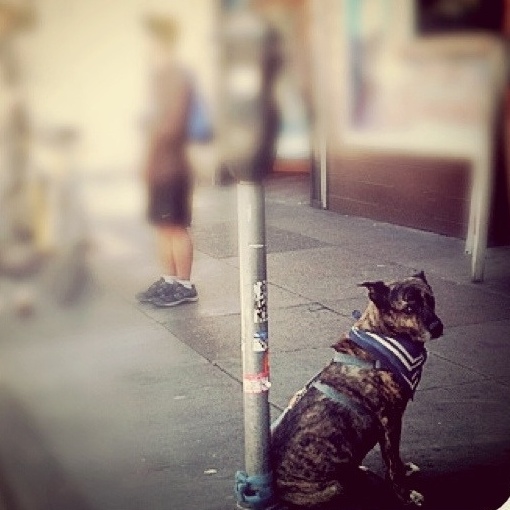} & \includegraphics[width=\linewidth,height=1.5cm,keepaspectratio]{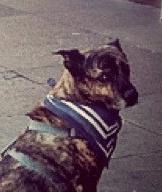} & \includegraphics[width=\linewidth,height=1.5cm,keepaspectratio]{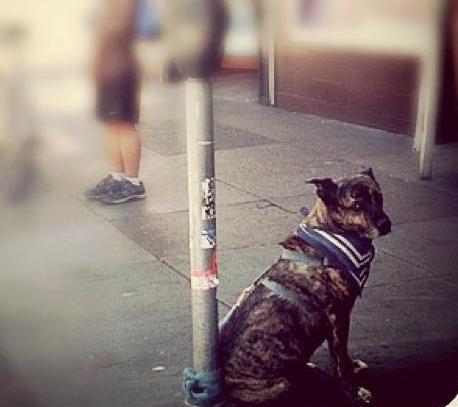} & \includegraphics[width=\linewidth,height=1.5cm,keepaspectratio]{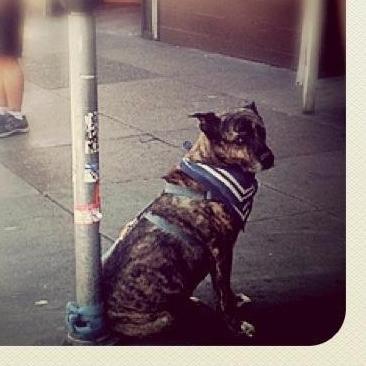} & \includegraphics[width=\linewidth,height=1.5cm,keepaspectratio]{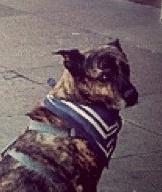} & \includegraphics[width=\linewidth,height=1.5cm,keepaspectratio]{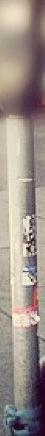} & \includegraphics[width=\linewidth,height=1.5cm,keepaspectratio]{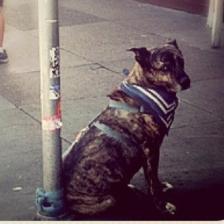}\\

       \includegraphics[width=\linewidth,height=1.5cm,keepaspectratio]{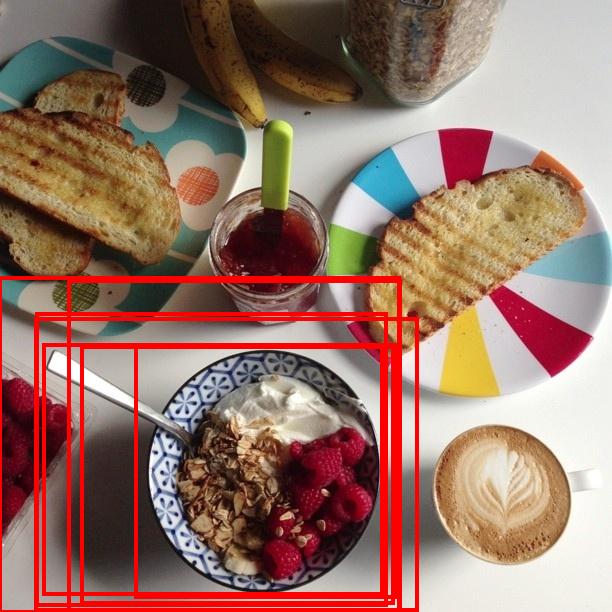} & 
       \vspace{-1.7cm}\scriptsize A blue bowl of white cream and red berries with a metal spoon in it & 
       \includegraphics[width=\linewidth,height=1.5cm,keepaspectratio]{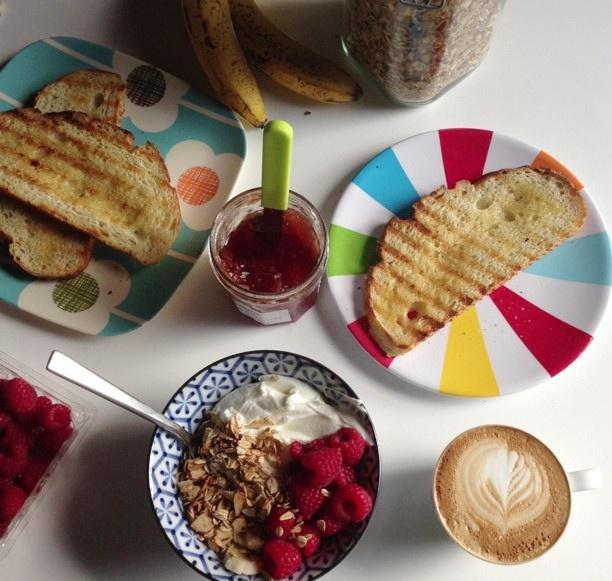} & \includegraphics[width=\linewidth,height=1.5cm,keepaspectratio]{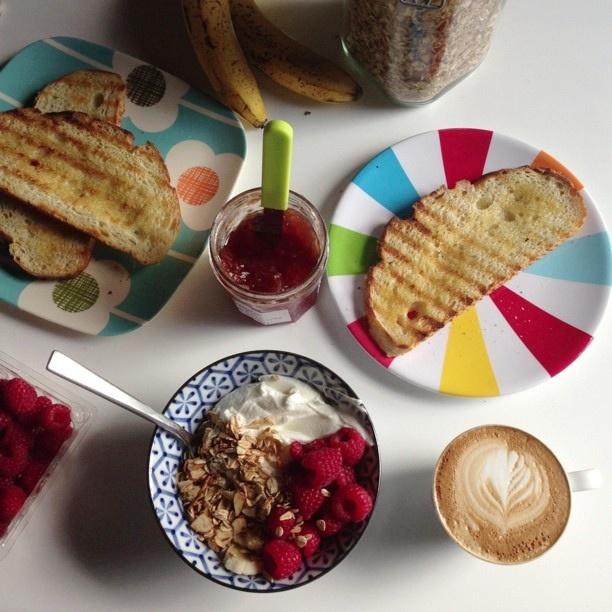} & \includegraphics[width=\linewidth,height=1.5cm,keepaspectratio]{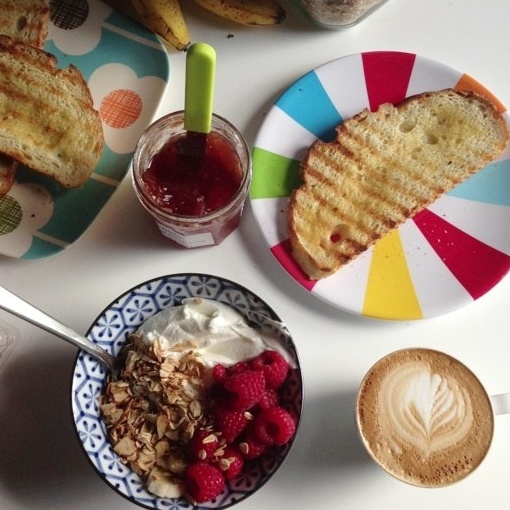} & \includegraphics[width=\linewidth,height=1.5cm,keepaspectratio]{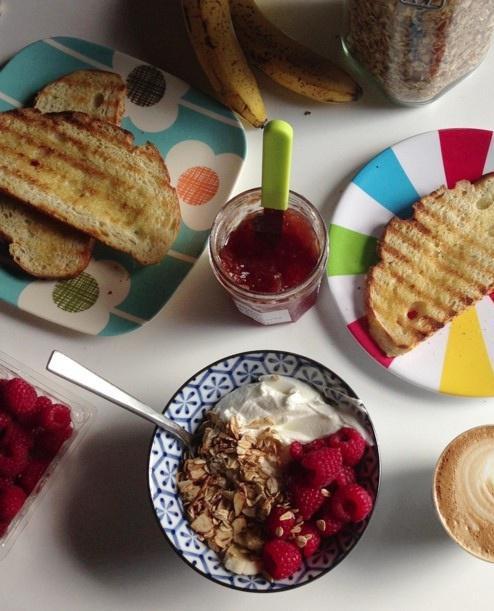} & \includegraphics[width=\linewidth,height=1.5cm,keepaspectratio]{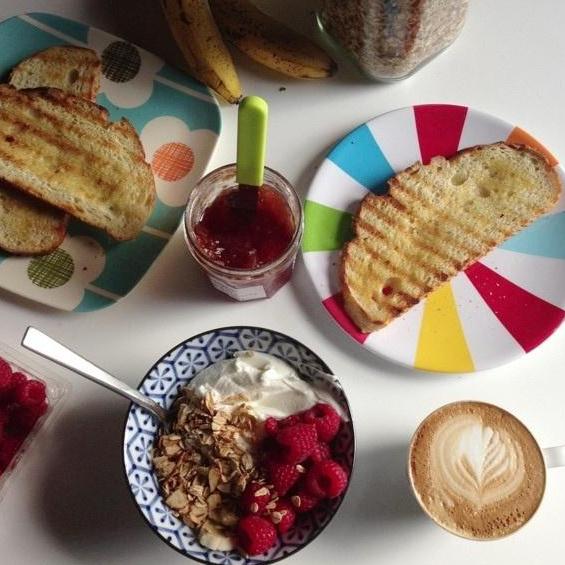} & \includegraphics[width=\linewidth,height=1.5cm,keepaspectratio]{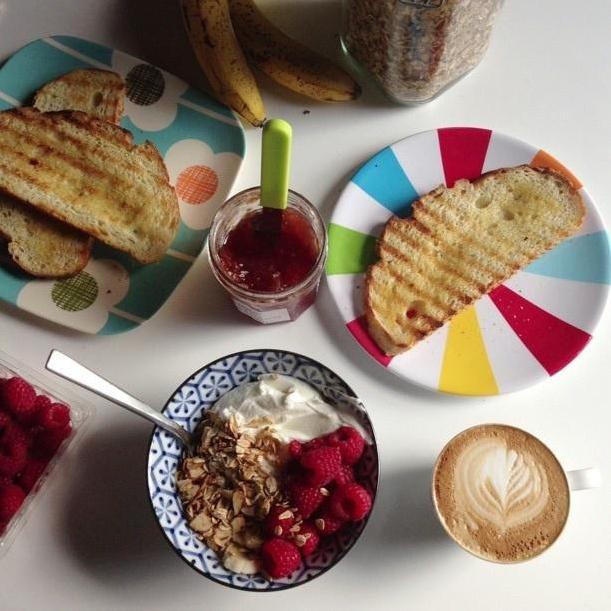} & \includegraphics[width=\linewidth,height=1.5cm,keepaspectratio]{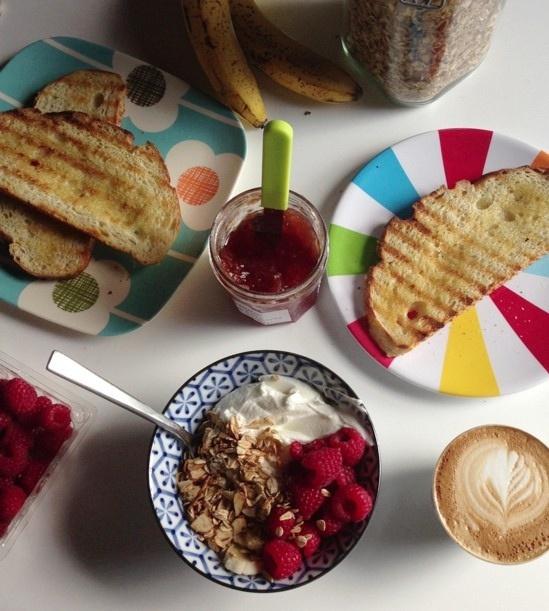} & \includegraphics[width=\linewidth,height=1.5cm,keepaspectratio]{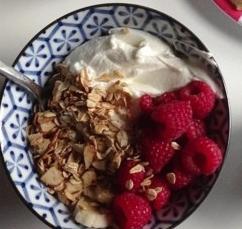} & \includegraphics[width=\linewidth,height=1.5cm,keepaspectratio]{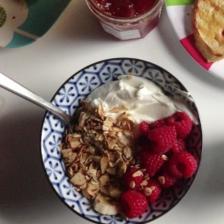}\\

       \includegraphics[width=\linewidth,height=1.5cm,keepaspectratio]{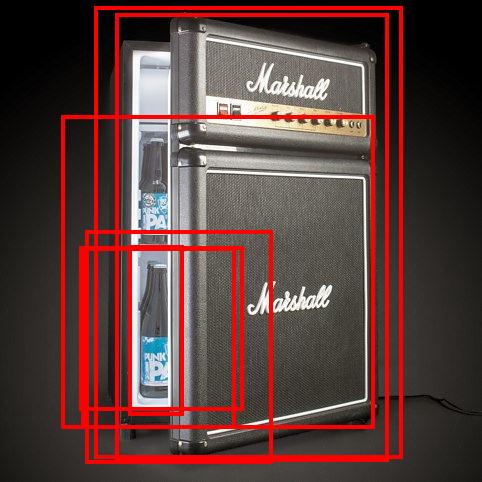} & 
       \vspace{-1.5cm}\scriptsize A bottle of juice is in the freezer & 
       \includegraphics[width=\linewidth,height=1.5cm,keepaspectratio]{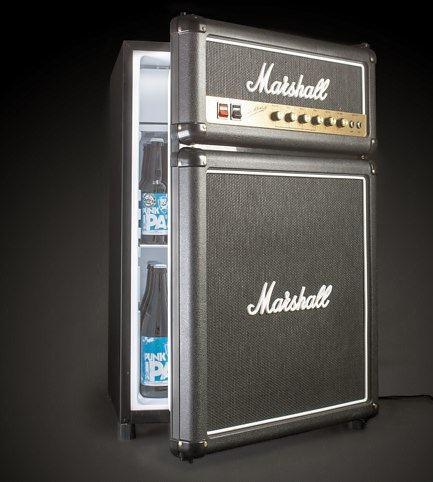} & \includegraphics[width=\linewidth,height=1.5cm,keepaspectratio]{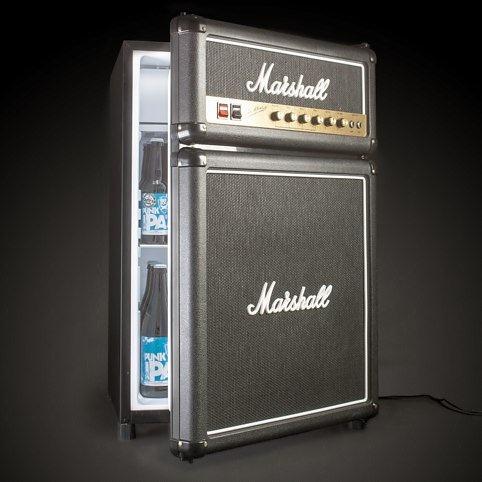} & \includegraphics[width=\linewidth,height=1.5cm,keepaspectratio]{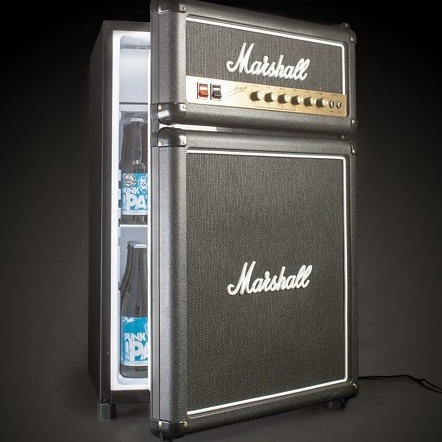} & \includegraphics[width=\linewidth,height=1.5cm,keepaspectratio]{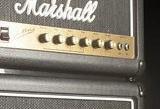} & \includegraphics[width=\linewidth,height=1.5cm,keepaspectratio]{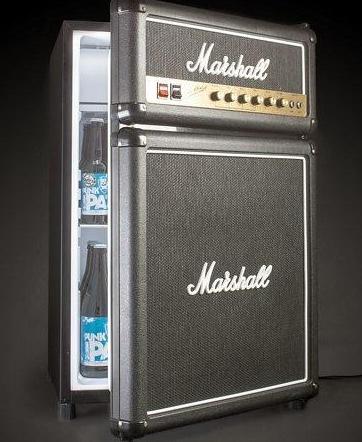} & \includegraphics[width=\linewidth,height=1.5cm,keepaspectratio]{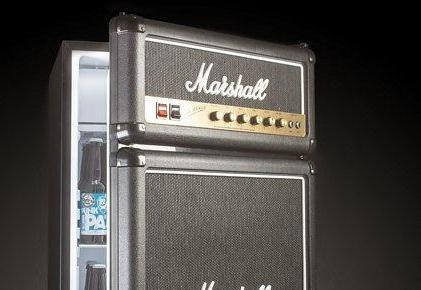} & \includegraphics[width=\linewidth,height=1.5cm,keepaspectratio]{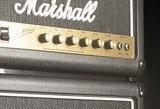} & \includegraphics[width=\linewidth,height=1.5cm,keepaspectratio]{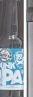} & \includegraphics[width=\linewidth,height=1.5cm,keepaspectratio]{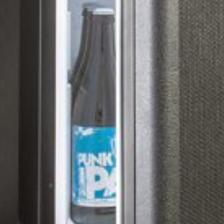} \\        
       \bottomrule
     \end{tabular}
     }
     \caption{The cropped images obtained by the proposed method and the eight baseline methods. The user defined ground truth bounding box annotations are shown on the original images with red. The proposed method well crops the images as user described.
}
   \label{fig:qualitative}
\end{figure*}

%% file: tbl/quantitative/item.tex
\newcolumntype{C}{>{\centering\arraybackslash}X}
\begin{table}
\small
\begin{center}
\resizebox{0.9\linewidth}{!}{%
\begin{tabularx}{\linewidth}{r | C}
\toprule
Method & Mean $\pm$ Std.  \\ 
\midrule
Original & 0.2869 $\pm $ 0.0280 \\
A2-RL\cite{li2018a2}  & 0.2975 $\pm $ 0.0238 \\
VPN\cite{goodview2018} & 0.3151 $\pm $ 0.0233\\
Anchor \cite{Zeng_2019_CVPR} & 0.3325 $\pm$ 0.0236  \\
GradCAM+A2-RL  & 0.3465 $\pm $ 0.0108  \\
GradCAM+Anchor & 0.3554 $\pm$ 0.0199 \\
GradCAM+VPN &0.3561 $\pm $ 0.0126 \\
GradCAM\cite{Selvaraju2017GradCAMVE} & 0.3597 $\pm$ 0.2017  \\
MAttNet\cite{mattnet}  & 0.3851 $\pm $ 0.2607   \\ 
\midrule
CAGIC  & \textbf{0.4160 $\pm $ 0.0129} \\

\bottomrule
\end{tabularx}%
}
\caption{
Quantitative comparison of the different methods using IoU measure on the output bounding boxes.}
\label{tbl:overlap}
\end{center}
\end{table}

%% file: tbl/quantitative/item2.tex
\newcolumntype{C}{>{\centering\arraybackslash}X}
\begin{table*}[t]
\small
\begin{center}
\resizebox{\linewidth}{!}{
\begin{tabularx}{\linewidth}{r | C C C C C C C}
\toprule
Method & Blue-1 & Blue-2 & Blue-3 & Blue-4 & METEOR & ROUGE\_L & CIDEr  \\ 
\midrule
GradCAM\cite{Selvaraju2017GradCAMVE} 
& 0.2728 
&   0.1410
&  0.0874
&   0.0531
&  0.1182
&  0.2790
&  0.6973 \\
MAttNet\cite{mattnet}  
& 0.1718
& 0.0937
& 0.0603
& 0.0355
& 0.1132
& 0.2947
& 0.7154  \\ 
\midrule
CAGIC 
& \textbf{0.3424}
& \textbf{0.1876}
& \textbf{0.1017}
& \textbf{0.0631}
& \textbf{0.1702}
& \textbf{0.2970}
& \textbf{0.9054} \\\bottomrule
\end{tabularx}}

\caption{
Comparison on user intention presence. We ask users to caption cropped images and compare with natural language metrics how similar they are with the original desired caption.
}
\label{tbl:crosscrop}
\end{center}
\end{table*}

%% file: tbl/user/item.tex
\newcolumntype{C}{>{\centering\arraybackslash}X}
\begin{table}[t!]
\begin{center}
\resizebox{\linewidth}{!}{

\begin{tabularx}{\linewidth}{P{1.6cm} | C | C | C | C}
\toprule
 &\small{Original Image} & \small{MAttNet\cite{mattnet}}  & \small{GradCAM\cite{Selvaraju2017GradCAMVE}} &     \small{CAGIC} \\ \midrule
\scriptsize{Aggregated percentage $(\%)$ }& \vspace{0.005cm}21.04 & \vspace{0.005cm}23.93 & \vspace{0.005cm}25.51 & \vspace{0.005cm}\textbf{29.52}\\
\bottomrule
\end{tabularx}
}

\caption{
Top three methods of the qualitative comparison along with original image were compared through human survey and evaluated by aggregation.}
\label{tbl:user}
\end{center}

\end{table}

%% file: fig/ablation/item.tex
\begin{figure*}[t]
\centering
\includegraphics[width=\linewidth]{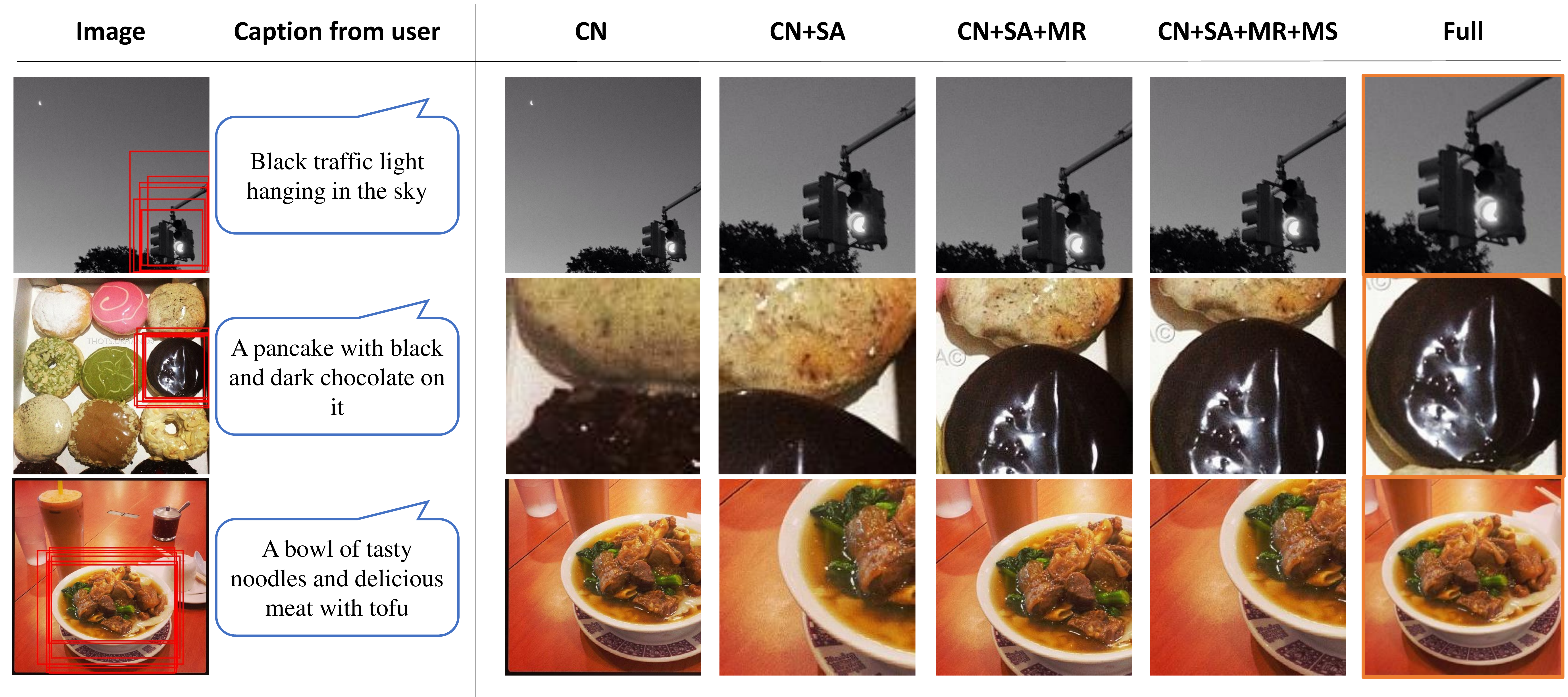}
\caption{Qualitative comparison between our methods which consist of the combination of Caption Network (CN), Scale Anneal (SA), Multiple Restart (MR) and Multi-Scale Bilinear Sampling (MS). The results show that the combination of all of these elements (proposed full method) together can provide the best output image.} 
\label{fig:ablation} 
\end{figure*}

%% file: fig/Aesthetics_ablation/item.tex
\begin{figure}[t!]
\centering
\includegraphics[width=1\linewidth]{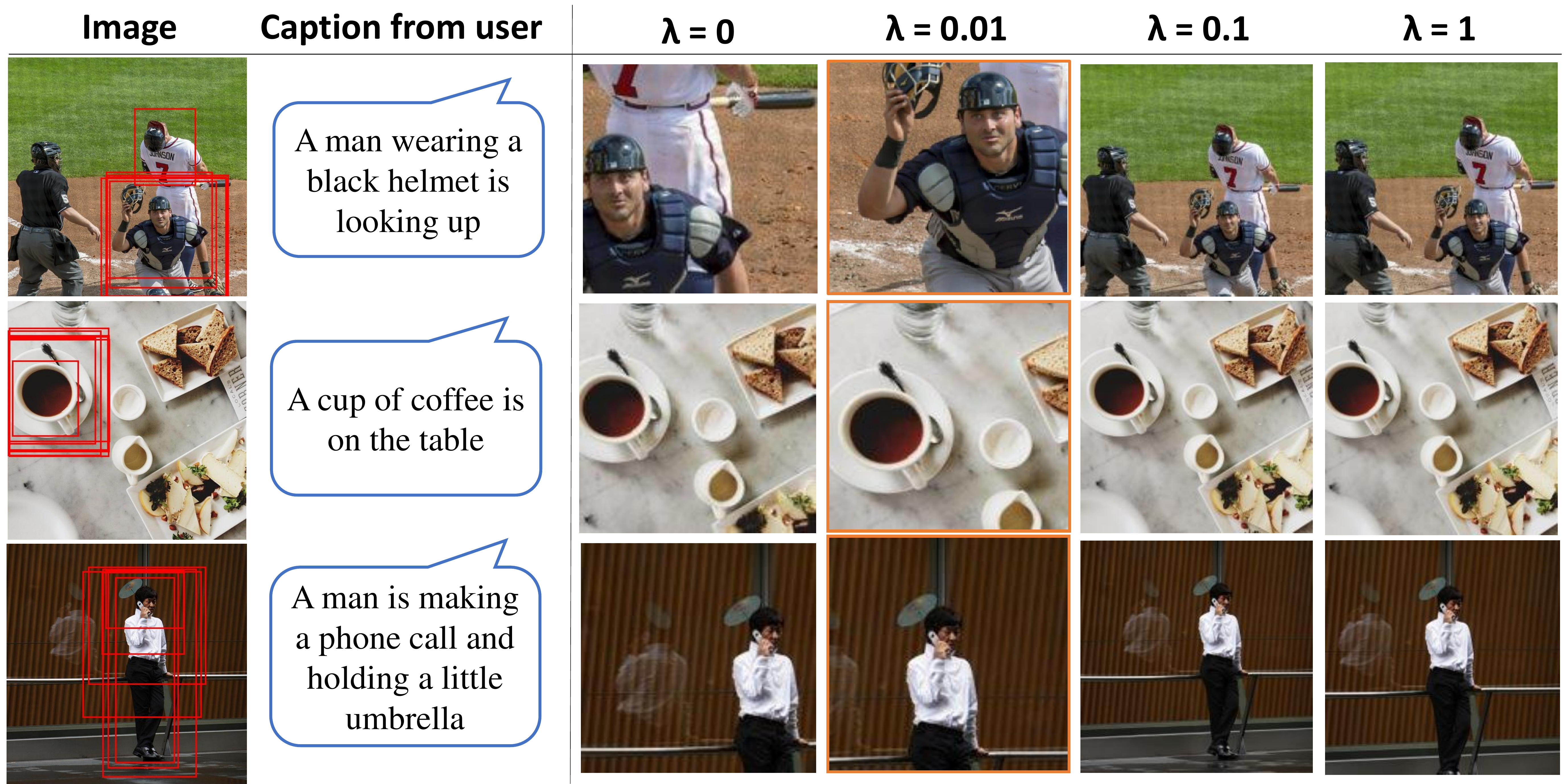}
\caption{Comparison of different aesthetics ratios.}
\label{fig:aesth_ablation}
\end{figure}

%% file: tbl/ablation/item_quantitative.tex
\newcolumntype{C}{>{\centering\arraybackslash}X}
\begin{table}
\small
\begin{center}
\resizebox{0.9\linewidth}{!}{%
\begin{tabularx}{\linewidth}{r | C}
\toprule
Method & Mean $\pm$ Std.  \\ 
\midrule
CN & 0.3338 $\pm $ 0.0693 \\
CN+SA & 0.3406 $\pm $ 0.0584 \\
CN+SA+MR & 0.3562 $\pm $ 0.0565\\
CN+SA+MR+MS  & 0.3816 $\pm$ 0.0615  \\ \midrule
CAGIC (Full)  & \textbf{0.4160} $\pm $ \textbf{0.0129}  \\
\bottomrule
\end{tabularx}%
}
\caption{
Quantitative result of the ablation study using IoU measure on the output bounding boxes}
\label{tbl:ablationquantitative}
\end{center}
\end{table}

%% file: tex/conclusion.tex
\section{Conclusion}

In this paper, we have proposed a novel optimization framework that produces image crops that follow users' descriptions and aesthetics criteria. The main idea behind this study was to demonstrate that this can successfully be achieved without training a specialized network, but instead utilizing two pre-trained networks on related tasks, namely image captioning and aesthetics measuring. 
We designed a new cost function considering the two networks' outputs together and performed a search among various bilinear sampling parameters.
However, the parameter space is very large and non-convex, so to achieve a stable and efficient solution, we designed a new scale annealing and multiple restarts search strategy. 
We have shown that our proposed method outperforms other existing approaches which are based on either solely saliency-based or caption-based cropping methods.

Our ongoing research is extending in two directions. The description- and aesthetics-driven captioning will further be coupled with attention mechanisms and gaze estimation.
On a more conceptual level, we are extending the ideas of solving complex tasks in optimization frameworks by repurposing existing modular networks trained on auxiliary tasks, thereby seeking efficient alternatives to extensive learning of networks from scratch.

%% file: tex/acknowledge.tex
\section*{Acknowledgements}
This work was partially supported by the Natural Sciences and Engineering
Research Council of Canada (NSERC) Discovery Grant ``Deep Visual Geometry
Machines'' (RGPIN-2018-03788), and by systems supplied by Compute Canada.
We acknowledge MoD/Dstl and EPSRC for providing the grant to support the
UK academics involvement in a Department of Defense funded MURI project
through EPSRC grant EP/N019415/1, and IITP grant funded by the Korea government (MSIT) (IITP-2020-0-01789, ITRC support program). We thank NVIDIA for sponsoring one GeForce Titan Xp GPU. This work was supported by the Institute of Information and communications Technology Planning and evaluation (IITP) grant funded by the Korean government (MSIT) (2021-0-00537, Visual common sense through self-supervised learning for restoration of invisible parts in images).

%% file: main.bbl
\begin{thebibliography}{10}
\expandafter\ifx\csname url\endcsname\relax
  \def\url#1{\texttt{#1}}\fi
\expandafter\ifx\csname urlprefix\endcsname\relax\def\urlprefix{URL }\fi
\expandafter\ifx\csname href\endcsname\relax
  \def\href#1#2{#2} \def\path#1{#1}\fi

\bibitem{Huang2007GeneratingGT}
W.~Huang, C.~L. Tan, J.~Zhao, Generating ground truthed dataset of chart
  images: Automatic or semi-automatic?, in: GREC, 2007.

\bibitem{iswanto2017visual}
I.~A. Iswanto, B.~Li, Visual object tracking based on mean-shift and
  particle-kalman filter, Procedia computer science 116 (2017) 587--595.

\bibitem{chu2014optimized}
W.-T. Chu, C.-H. Yu, H.-H. Wang, Optimized comics-based storytelling for
  temporal image sequences, IEEE Transactions on Multimedia 17~(2) (2014)
  201--215.

\bibitem{ChenCVPR2016}
J.~Chen, G.~Bai, S.~Liang, Z.~Li, Automatic image cropping: A computational
  complexity study, 2016, pp. 507--515.
\newblock \href {https://doi.org/10.1109/CVPR.2016.61}
  {\path{doi:10.1109/CVPR.2016.61}}.

\bibitem{kao2017automatic}
Y.~Kao, R.~He, K.~Huang, Automatic image cropping with aesthetic map and
  gradient energy map, IEEE, 2017, pp. 1982--1986.

\bibitem{cornia2018automatic}
M.~Cornia, S.~Pini, L.~Baraldi, R.~Cucchiara, Automatic image cropping and
  selection using saliency: An application to historical manuscripts, in:
  Italian Research Conference on Digital Libraries, Springer, 2018, pp.
  169--179.

\bibitem{guo2018}
G.~Guo, H.~Wang, C.~Shen, Y.~Yan, H.~M. Liao, Automatic image cropping for
  visual aesthetic enhancement using deep neural networks and cascaded
  regression, IEEE Transactions on Multimedia 20~(8) (2018) 2073--2085.
\newblock \href {https://doi.org/10.1109/TMM.2018.2794262}
  {\path{doi:10.1109/TMM.2018.2794262}}.

\bibitem{shan2018photobomb}
N.~Shan, D.~S. Tan, M.~S. Denekew, Y.-Y. Chen, W.-H. Cheng, K.-L. Hua,
  Photobomb defusal expert: Automatically remove distracting people from
  photos, IEEE Transactions on Emerging Topics in Computational
  Intelligence~(99) (2018) 1--11.

\bibitem{li2018a2}
D.~Li, H.~Wu, J.~Zhang, K.~Huang, {A2-RL: Aesthetics Aware Reinforcement
  Learning for Image Cropping}, 2018, pp. 8193--8201.

\bibitem{chen2017quantitative}
Y.-L. Chen, T.-W. Huang, K.-H. Chang, Y.-C. Tsai, H.-T. Chen, B.-Y. Chen,
  Quantitative analysis of automatic image cropping algorithms: A dataset and
  comparative study, IEEE, 2017, pp. 226--234.

\bibitem{ChenACM2017}
Y.-L. Chen, J.~Klopp, M.~Sun, S.-Y. Chien, K.-L. Ma, Learning to compose with
  professional photographs on the web, in: ACM International Conference on
  Multimedia, ACM, 2017, pp. 37--45.

\bibitem{mattnet}
L.~Yu, Z.~Lin, X.~Shen, J.~Yang, X.~Lu, M.~Bansal, T.~L. Berg, Mattnet: Modular
  attention network for referring expression comprehension, in: CVPR, 2018, pp.
  1307--1315.

\bibitem{grounding2016}
A.~Rohrbach, M.~Rohrbach, R.~Hu, T.~Darrell, B.~Schiele, Grounding of textual
  phrases in images by reconstruction, in: ECCV, Springer, 2016, pp. 817--834.

\bibitem{Jaderberg15}
M.~Jaderberg, K.~Simonyan, A.~Zisserman, K.~Kavukcuoglu, Spatial transformer
  networks, in: NIPS, 2015.

\bibitem{xu2015show}
K.~Xu, J.~Ba, R.~Kiros, K.~Cho, A.~Courville, R.~Salakhudinov, R.~Zemel,
  Y.~Bengio, Show, attend and tell: Neural image caption generation with visual
  attention, in: International conference on machine learning, 2015, pp.
  2048--2057.

\bibitem{huang2015automatic}
J.~Huang, H.~Chen, B.~Wang, S.~Lin, Automatic thumbnail generation based on
  visual representativeness and foreground recognizability, 2015, pp. 253--261.

\bibitem{jaiswal2015saliency}
N.~Jaiswal, Y.~K. Meghrajani, Saliency based automatic image cropping using
  support vector machine classifier, in: ICIIECS, IEEE, 2015, pp. 1--5.

\bibitem{choi2016object}
J.~Choi, C.~Kim, Object-aware image thumbnailing using image classification and
  enhanced detection of roi, Multimedia Tools and Applications 75~(23) (2016)
  16191--16207.

\bibitem{FangACM2014}
C.~Fang, Z.~Lin, R.~Mech, X.~Shen, Automatic image cropping using visual
  composition, boundary simplicity and content preservation models, in: ACM
  international conference on Multimedia, ACM, 2014, pp. 1105--1108.

\bibitem{wang2015saliency}
W.~Wang, J.~Shen, F.~Porikli, Saliency-aware geodesic video object
  segmentation, 2015, pp. 3395--3402.

\bibitem{wang2015consistent}
W.~Wang, J.~Shen, L.~Shao, Consistent video saliency using local gradient flow
  optimization and global refinement, IEEE Transactions on Imnage Processing
  24~(11) (2015) 4185--4196.

\bibitem{wang2016correspondence}
W.~Wang, J.~Shen, L.~Shao, F.~Porikli, Correspondence driven saliency transfer
  25~(11) (2016) 5025--5034.

\bibitem{wang2018saliency}
W.~Wang, J.~Shen, R.~Yang, F.~Porikli, Saliency-aware video object
  segmentation~(1) (2018) 20--33.

\bibitem{islam2017survey}
M.~B. Islam, W.~Lai-Kuan, W.~Chee-Onn, A survey of aesthetics-driven image
  recomposition, Multimedia Tools and Applications 76~(7) (2017) 9517--9542.

\bibitem{lu2019aesthetic}
P.~Lu, H.~Zhang, X.~Peng, X.~Peng, Aesthetic guided deep regression network for
  image cropping, Signal Processing: Image Communication (2019).

\bibitem{hong2017cnn}
E.~Hong, J.~Jeon, S.~Lee, Cnn based repeated cropping for photo composition
  enhancement, 2017.

\bibitem{wang2015learning}
P.~Wang, Z.~Lin, R.~Mech, Learning an aesthetic photo cropping cascade, IEEE,
  2015, pp. 448--455.

\bibitem{zhang2014weakly}
L.~Zhang, M.~Song, Y.~Yang, Q.~Zhao, C.~Zhao, N.~Sebe, Weakly supervised photo
  cropping, IEEE Transactions on Multimedia 16~(1) (2014) 94--107.

\bibitem{goodview2018}
Z.~Wei, J.~Zhang, X.~Shen, Z.~Lin, R.~Mech, M.~Hoai, D.~Samaras, Good view
  hunting: learning photo composition from dense view pairs, in: CVPR, 2018,
  pp. 5437--5446.

\bibitem{datta2019align2ground}
S.~Datta, K.~Sikka, A.~Roy, K.~Ahuja, D.~Parikh, A.~Divakaran, Align2ground:
  Weakly supervised phrase grounding guided by image-caption alignment, in:
  Proceedings of the IEEE International Conference on Computer Vision, 2019,
  pp. 2601--2610.

\bibitem{bai2018survey}
S.~Bai, S.~An, A survey on automatic image caption generation, Neurocomputing
  311 (2018) 291--304.

\bibitem{Lin2014MicrosoftCC}
T.-Y. Lin, M.~Maire, S.~J. Belongie, L.~D. Bourdev, R.~B. Girshick, J.~Hays,
  P.~Perona, D.~Ramanan, P.~Doll{\'a}r, C.~L. Zitnick, Microsoft coco: Common
  objects in context, 2014.

\bibitem{multistart}
G.~Song, B.~B. Avants, J.~C. Gee, Multi-start method with prior learning for
  image registration, 2007, pp. 1--8.
\newblock \href {https://doi.org/10.1109/ICCV.2007.4409159}
  {\path{doi:10.1109/ICCV.2007.4409159}}.

\bibitem{kwedlo2015new}
W.~Kwedlo, A new random approach for initialization of the multiple restart em
  algorithm for gaussian model-based clustering, Pattern Analysis and
  Applications 18~(4) (2015) 757--770.

\bibitem{byrd1995limited}
R.~H. Byrd, P.~Lu, J.~Nocedal, C.~Zhu, A limited memory algorithm for bound
  constrained optimization, SIAM Journal on Scientific Computing 16~(5) (1995)
  1190--1208.

\bibitem{Tensorflow}
M.~Abadi, P.~Barham, J.~Chen, Z.~Chen, A.~Davis, J.~Dean, M.~Devin,
  S.~Ghemawat, G.~Irving, M.~Isard, M.~Kudlur, J.~Levenberg, R.~Monga,
  S.~Moore, D.~Murray, B.~Steiner, P.~Tucker, V.~Vasudevan, P.~Warden,
  M.~Wicke, Y.~Yu, X.~Zheng, {Tensorflow: A System for Large-Scale Machine
  Learning}, in: USENIX Conference on Operating Systems Design and
  Implementation, 2016, pp. 265--283.

\bibitem{kazemzadeh2014referitgame}
S.~Kazemzadeh, V.~Ordonez, M.~Matten, T.~Berg, Referitgame: Referring to
  objects in photographs of natural scenes, in: EMNLP, 2014, pp. 787--798.

\bibitem{krishnavisualgenome}
R.~Krishna, Y.~Zhu, O.~Groth, J.~Johnson, K.~Hata, J.~Kravitz, S.~Chen,
  Y.~Kalantidis, L.-J. Li, D.~A. Shamma, et~al., Visual genome: Connecting
  language and vision using crowdsourced dense image annotations, International
  journal of computer vision 123~(1) (2017) 32--73.

\bibitem{thomee2015yfcc100m}
B.~Thomee, D.~A. Shamma, G.~Friedland, B.~Elizalde, K.~Ni, D.~Poland, D.~Borth,
  L.-J. Li, Yfcc100m: The new data in multimedia research, Communications of
  the ACM 59~(2) (2016) 64--73.

\bibitem{overlap}
M.~Kristan, J.~Matas, A.~Leonardis, T.~Voj{\'\i}{\v{r}}, R.~Pflugfelder,
  G.~Fernandez, G.~Nebehay, F.~Porikli, L.~{\v{C}}ehovin, A novel performance
  evaluation methodology for single-target trackers, IEEE transactions on PAMI
  38~(11) (2016) 2137--2155.

\bibitem{Selvaraju2017GradCAMVE}
R.~R. Selvaraju, M.~Cogswell, A.~Das, R.~Vedantam, D.~Parikh, D.~Batra,
  Grad-cam: Visual explanations from deep networks via gradient-based
  localization (2017) 618--626.

\bibitem{Zeng_2019_CVPR}
H.~Zeng, L.~Li, Z.~Cao, L.~Zhang, Reliable and efficient image cropping: A grid
  anchor based approach, in: The IEEE Conference on Computer Vision and Pattern
  Recognition (CVPR), 2019.

\bibitem{Gatys16a}
L.~A. Gatys, A.~S. Ecker, M.~Bethge, {Image Style Transfer Using Convolutional
  Neural Networks}, 2016.

\bibitem{Johnson16}
J.~Johnson, A.~Alahi, L.~Fei{-}fei, {Perceptual Losses for Real-Time Style
  Transfer and Super-Resolution}, 2016, pp. 694--711.

\end{thebibliography}
